\documentclass[twoside,11pt]{article}

\usepackage{blindtext}

\usepackage[preprint]{jmlr2e}

\usepackage{lastpage}
\jmlrheading{--}{2024}{1-\pageref{LastPage}}{4/24}{--}{--}{Masaki Adachi, Satoshi Hayakawa, Martin Jørgensen, Saad Hamid, Harald Oberhauser, Michael A. Osborne}

\usepackage{amsthm}
\usepackage{amsmath}
\usepackage{pifont}              
\usepackage{caption}             
\usepackage{booktabs}            
\usepackage{xcolor}              
\usepackage{mathtools}           
\usepackage{comment}             
\usepackage{paralist}            
\usepackage{tcolorbox}           
\usepackage{subfig}              
\usepackage{algorithm}           
\usepackage{algorithmic}         
\newcommand{\xmark}{\textcolor{gray!30}{\ding{55}}}

\makeatletter
\def\blfootnote{\xdef\@thefnmark{*}\@footnotetext}
\makeatother

\usepackage{amsmath,amsfonts,bm}

\def\eqref#1{equation~\ref{#1}}

\def\1{\bm{1}}

\usepackage{thmtools}
\usepackage{thm-restate}
\usepackage{amsmath}
\newtheorem{defn}{Definition}

\newtheorem{theorem}[defn]{Theorem}

\newtheorem{prop}{Proposition}[]
\DeclareMathOperator*{\plim}{plim}

\ShortHeadings{SOBER}{Adachi, Hayakawa, Jørgensen, Hamid, Oberhauser, and Osborne}
\firstpageno{1}

\begin{document}

\title{A Quadrature Approach for General-Purpose\\ Batch Bayesian Optimization via Probabilistic Lifting}

\author{\name Masaki Adachi \email masaki@robots.ox.ac.uk\\
       \addr Machine Learning Research Group, University of Oxford, UK
       \AND
       \name Satoshi Hayakawa \email hayakawa@maths.ox.ac.uk\\
       \addr Mathematical Institute, University of Oxford, UK
       \AND
       \name Martin Jørgensen \email  martin.jorgensen@helsinki.fi\\
       \addr Department of Computer Science, University of Helsinki, Finland
       \AND
       \name Saad Hamid\footnote{Work done while at Machine Learning Research Group, University of Oxford, UK} \email  saad.hamid@aioilab-oxford.eu\\
       \addr Aioi R\&D Lab, Oxford, UK, and Mind Foundry, Oxford, UK
       \AND
       \name Harald Oberhauser \email oberhauser@maths.ox.ac.uk\\
       \addr Mathematical Institute, University of Oxford, UK
       \AND
       \name Michael A. Osborne \email  mosb@robots.ox.ac.uk\\
       \addr Machine Learning Research Group, University of Oxford, UK
       }
\editor{My editor}

\maketitle

\begin{abstract}

Parallelisation in Bayesian optimisation is a common strategy but faces several challenges: the need for flexibility in acquisition functions and kernel choices, flexibility dealing with discrete and continuous variables simultaneously, model misspecification, and lastly fast massive parallelisation. To address these challenges, we introduce a versatile and modular framework for batch Bayesian optimisation via probabilistic lifting with kernel quadrature, called \emph{SOBER}, which we present as a Python library based on GPyTorch/BoTorch. Our framework offers the following unique benefits: (1) Versatility in downstream tasks under a unified approach. (2) A gradient-free sampler, which does not require the gradient of acquisition functions, offering domain-agnostic sampling (e.g., discrete and mixed variables, non-Euclidean space). (3) Flexibility in domain prior distribution. (4) Adaptive batch size (autonomous determination of the optimal batch size). (5) Robustness against a misspecified reproducing kernel Hilbert space. (6) Natural stopping criterion.

\end{abstract}

\begin{keywords}
  Batch Bayesian Optimisation, Bayesian Quadrature, Kernel Quadrature
\end{keywords}

\blfootnote{Work done while at Machine Learning Research Group, University of Oxford, UK}

\section{Introduction}
\begin{figure*}[hbt!]
  \centering
  \includegraphics[width=1\hsize]{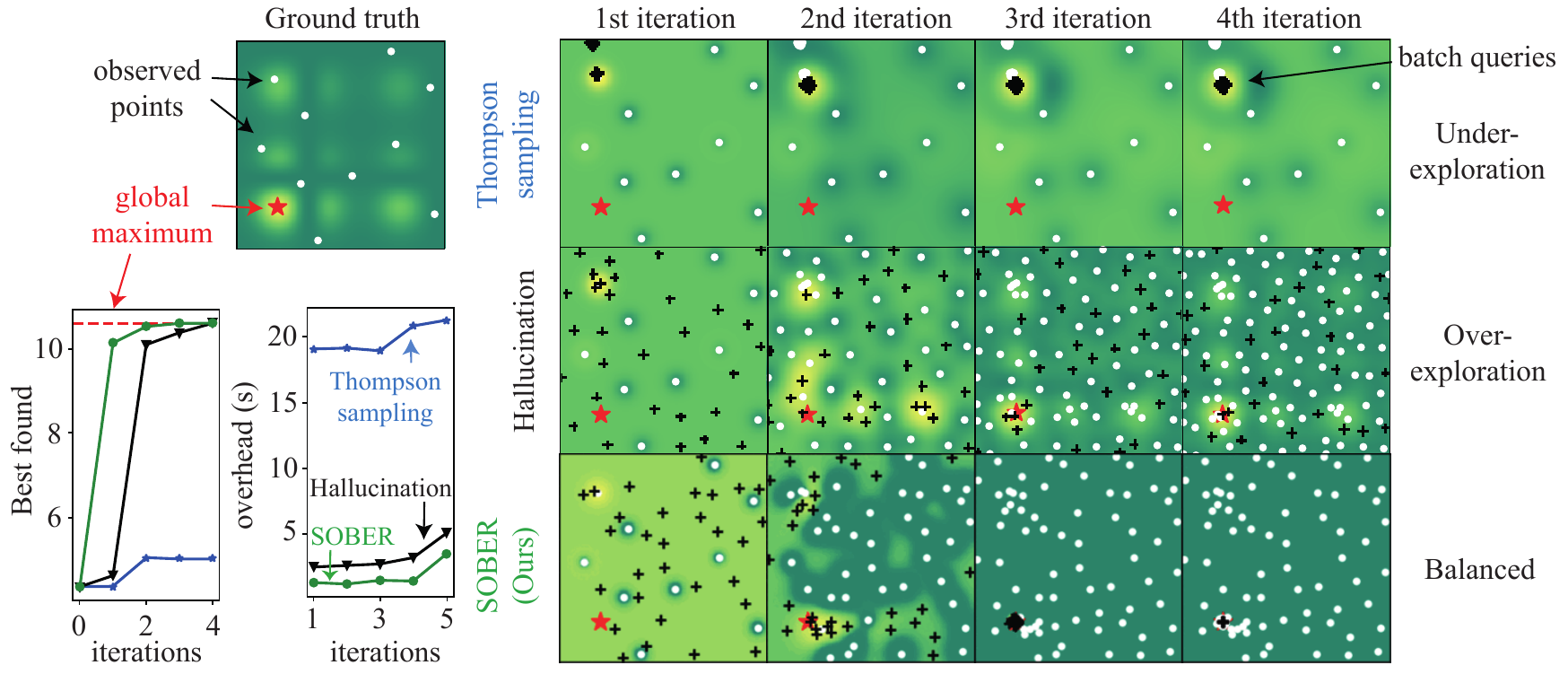}
  \caption{
  A demonstrating example featuring 2D Branin-Hoo function with nine peaks and the global maximum at the bottom-left corner (red star). Initial 10 i.i.d. samples (white dots) unluckily misidentify the top-left peak as the promising area. Thompson sampling (blue lines) under-explores, erroneously focusing 30 queries (black crosses) near the top-left. Conversely, hallucination (black lines) over-explores, constantly venturing into new regions, yet allocating only a few queries towards the bottom-left area. Our SOBER approach (green lines) starts with wide exploration, then narrows down to the global maximum, demonstrating balanced exploration. The convergence plot illustrates that SOBER outperforms the baselines with the least wall-clock time overhead. The image’s colour scheme represents different functions: upper confidence bound for Thompson and hallucination, log$\pi$ for SOBER.
  }
  \label{fig:demo}
  \vspace{-1em}
\end{figure*}
\begin{table}[ht]
    \centering
    \resizebox{1\textwidth}{!}{
    \begin{tabular}{lccccccccccc}
    \toprule
         Batch methods &
         task&
         \begin{tabular}{@{}l@{}}gradient- \\ free?\end{tabular}&
         \begin{tabular}{@{}l@{}}adaptive \\ batch?\end{tabular}&
         \begin{tabular}{@{}l@{}}misspec. \\ RKHS?\end{tabular}&
         \begin{tabular}{@{}l@{}}stopping \\ criterion?\end{tabular}&
         \begin{tabular}{@{}l@{}}Any \\ AF?\end{tabular}&
         \begin{tabular}{@{}l@{}}Any \\ kernel?\end{tabular}&
         \begin{tabular}{@{}l@{}}const-\\raint?\end{tabular}&
         \begin{tabular}{@{}l@{}}large \\ batch?\end{tabular}&
         \begin{tabular}{@{}l@{}}discrete \\ space?\end{tabular}&
         \begin{tabular}{@{}l@{}}non- \\ Euclidean?\end{tabular}
         \\
    \midrule
         \begin{tabular}{@{}l@{}}Batch Thompson sampling\\ \citep{kandasamy2018parallelised} \end{tabular}& BO & \ding{51} & \xmark & \xmark& \xmark& \xmark &\ding{51} & \xmark & \ding{51} & \ding{51} & \ding{51} \\
         \begin{tabular}{@{}l@{}}DPP-TS\\ \citep{nava2022diversified} \end{tabular}&  BO & \ding{51} & \xmark & \xmark& \xmark& \xmark &\ding{51} & \xmark & \xmark & \ding{51} & \ding{51}\\
         \begin{tabular}{@{}l@{}}MC-SAA\\ \citep{balandat2020botorch} \end{tabular}& BO & \xmark& \xmark & \xmark& \xmark& \xmark &\ding{51} & \ding{51} & \ding{51} & \xmark & \ding{51}\\
         \begin{tabular}{@{}l@{}}TurBO\\ \citep{eriksson2019scalable} \end{tabular}& BO & \xmark & \xmark & \xmark& \xmark& \xmark & \xmark & \xmark & \ding{51} & \xmark & \xmark\\
         \begin{tabular}{@{}l@{}}SCBO\\ \citep{eriksson2021scalable} \end{tabular}&  BO & \ding{51} & \xmark & \xmark& \xmark& \xmark & \xmark & \ding{51} & \ding{51} & \ding{51} & \ding{51}\\
         \begin{tabular}{@{}l@{}}PESC\\ \citep{hernandez2015predictive} \end{tabular}& BO & \xmark & \xmark & \xmark& \xmark & \xmark& \ding{51} & \ding{51} & \xmark & \ding{51} & \ding{51} \\
         \begin{tabular}{@{}l@{}}GIBBON\\ \citep{moss2021gibbon} \end{tabular}& BO & \xmark & \xmark & \xmark& \xmark& \xmark & \ding{51} & \ding{51} & \xmark & \ding{51} & \ding{51}\\
         \begin{tabular}{@{}l@{}}B3O\\ \citep{nguyen2016budgeted} \end{tabular}& BO & \xmark & \ding{51} & \xmark& \xmark & \ding{51}& \ding{51} & \xmark & \xmark & \xmark & \xmark \\
         \begin{tabular}{@{}l@{}}Hallucination\\ \citep{azimi2010batch} \end{tabular}& Any & \xmark & \xmark & \xmark& \xmark & \ding{51}& \ding{51} & \ding{51} & \xmark & \ding{51} & \ding{51}\\
         \begin{tabular}{@{}l@{}}Local Penalisation\\ \citep{gonzalez2016batch} \end{tabular} & Any & \xmark & \xmark & \xmark& \xmark &\ding{51}& \ding{51} & \ding{51} & \xmark & \xmark & \xmark\\
    \midrule
         \textbf{SOBER (Ours)} & Any & \ding{51} & \ding{51} & \ding{51} & \ding{51} & \ding{51}& \ding{51} & \ding{51} & \ding{51} & \ding{51} & \ding{51}\\
    \bottomrule
    \end{tabular}
    }
    \caption{Comparisons between our proposed SOBER with popular batch methods. Task refers to batch BO, BQ, and AL. misspec. RKHS refers to bounded worst-case error against the misspeficied RKHS. Our SOBER is the most versatile algorithm with unique benefits.}
    \vspace{-1.5em}
    \label{tab:comparison}
\end{table}
Bayesian optimisation (BO; \citet{mockus1998application, garnett2023bayesian}) is a model-based global optimisation strategy for black-box functions. It involves constructing a surrogate model to approximate the function, and subsequently using the model to efficiently select forthcoming query points, thereby offering sample-efficient optimisation. However, with the advancement of the machine learning field, the complexity and variety of practical applications have increased. For example, many hyperparameter optimisation tasks involve mixed variables, diverging from the typical assumption of solely continuous or discrete variables \citep{ru2020bayesian, wan2021think, daulton2022bayesian}. Drug discovery \citep{gomez2018automatic, adachi2021high}, a prominent area for BO and experimental design, necessitates specialised kernels and non-Euclidean space due to the molecular and graph representations required \citep{griffiths2024gauche}. Furthermore, real-world tasks often operate under numerous constraints, and such constraint functions can be black-box (also known as unknown constraints, \citet{gelbart2014bayesian}). This complexity has spurred the development of numerous specialised acquisition functions (AFs). Furthermore, these AFs are often incompatible with each other, which is a hindrance for practitioners. Particularly in batch settings, where multiple points are selected simultaneously for parallelising the costly evaluations such as physical experiments, compatibility issues become more evident. The batch setting typically suffers from (1) no compatibility to arbitrary AFs, kernels, or downstream tasks (e.g., constrained optimisation), (2) limited scalability to large batch size, (3) under-/over-explorative samples. Figure \ref{fig:demo} demonstrates these issues in popular batch Thompson sampling (TS;  \citet{thompson1933likelihood, kandasamy2018parallelised}) and hallucination \citep{azimi2010batch}. Table \ref{tab:comparison} and §\ref{sec:related} delineates the details.

In response to these challenges, we introduce \emph{SOBER} (\underline{S}olving \underline{O}ptimisation as \underline{B}ayesian \underline{E}stimation via \underline{R}ecombination), which not only offers more balanced explorative sampling and faster computation times but also exhibits unique advantages: (1) adaptive batch sizes---autonomous determination of the optimal batch size at each iteration, (2) robustness against misspecified GPs---our worst-case error is uniformly bounded in misspecified reproducing Kernel Hilbert Spaces (RKHS), (3) stopping criterion as integral variance, and (4) the domain prior distribution---flexibility to model input domains based on any distribution, unlike typical uniform distribution assumptions.
This approach, as illustrated in Figure \ref{fig:demo} and summarised in Table \ref{tab:comparison}, positions our algorithm as a highly versatile solution not only for BO, but also for Active Learning (AL; \citet{settles.tr09}) and Bayesian Quadrature (BQ; \citet{o1991bayes, hennig2022probabilistic}). 
We firstly introduced the idea of \emph{probabilistic lifting} to batch BO, which enables us to leverage a flexible emph{kernel quadrature (KQ)} method, thereby offering a versatile and modular approach. As detailed later, specifying the downstream tasks, AFs, and variable types is equivalent to specification in domain distribution. Thus, users can enjoy a plug-and-play parallelisation library for AL, BO, and BQ interchangeably. We have created the open-source library SOBER based on PyTorch \citep{paszke2019pytorch}, GPyTorch \citep{gardner2018gpytorch}, and BoTorch \citep{balandat2020botorch}, providing detailed tutorials with varied use cases.

\begin{tcolorbox}[colback=white]
In summary, we offer:
\begin{compactenum}
    \item A modular and flexible open-sourced Python library for batch BO, AL, and BQ is ready for \texttt{pip install sober-bo} on \url{https://github.com/ma921/SOBER}, with versatility summarised in Table \ref{tab:comparison}.
    \item The unique benefits of adaptive batch sizes, robustness against misspecified RKHS, and domain prior distribution enhances further efficiency and flexibility.
    \item An evaluation of the performance of SOBER against baselines in various synthetic and real-world tasks involving large batch sizes, mixed variables, constraints, and non-Euclidean space.
\end{compactenum}
\end{tcolorbox}

\section{Background} \label{sec:background}
In this section, we first introduce the GP, then review the batch BO tasks and related work.

\subsection{Gaussian Process}
Let $(\Omega, \mathcal{F}, \mathbb{P})$ be a probability space and $\mathcal{X} \subseteq \mathbb{R}^{d}$ be the input domain. A GP \citep{stein1999interpolation, rasmussen2006gaussian} is a stochastic process $g: \mathcal{X} \times \Omega \rightarrow \mathbb{R}$, whose properties are captured by the mean function $m: \mathcal{X} \rightarrow \mathbb{R}$, $m(x) = \mathbb{E}[g(x, \cdot)]$ and covariance function $K: \mathcal{X} \times \mathcal{X} \rightarrow \mathbb{R}$, $K(x, x^\prime) = \mathbb{E}[(g(x, \cdot) - m(x))(g(x^\prime, \cdot) - m(x^\prime))]$. The covariance function is symmetric ($K(x,x^\prime)=K(x^\prime, x), \forall x, x^\prime \in \mathcal{X}$) and positive definite ($\forall t \in \mathbb{N}, \{a_i\}^t_{i=1} \in \mathbb{R}, \{x_i\}^t_{i=1} \subset \mathcal{X}, \sum^t_{i,j=1} a_i a_j K(x_i, x_j) \geq 0$). We refer to any function satisfying the above two properties as a kernel. A GP induces a probability measure over functions, and is capable of conditioning on data in closed form for conjugate likelihood cases. In the regression setting, we further assume the labels $y = f(x) + \epsilon$, where $f$ is the function to estimate, $\epsilon \sim \mathcal{N}(0, \sigma^2)$ is i.i.d. zero-mean Gaussian noise, and $\sigma^2$ is the noise variance. Given a labelled data set $\mathcal{D}_t = \{ x_i, y_i\}_{i=1}^t := (\textbf{X}_t, \textbf{Y}_t)$ and corresponding covariance matrix $\textbf{K}_{XX} = (K(x_i, x^\prime_j))_{1\leq i,j\leq t} \in \mathbb{R}^{t \times t}$, the conditioned GP regression model is given by $f \mid D_t \sim \mathcal{GP}(m_t, C_t)$, where
\begin{align*}
    \begin{split}
    m_t(x) &= m(x) + K(x, \textbf{X}_t) (\textbf{K}_{XX} + \sigma^2 \textbf{I}_{t \times t})^{-1}(\textbf{Y}_t -m(\textbf{X}_t)),\\
    C_t(x, x^\prime) &= K(x, x^\prime) - K(x, \textbf{X}_t) (\textbf{K}_{XX} + \sigma^2 \textbf{I}_{t \times t})^{-1} K(\textbf{X}_t, x^\prime),
    \end{split}
\end{align*}
$m_t(\cdot)$ and $C_t(\cdot, \cdot)$ are the mean and covariance functions of the GP posterior predictive distribution conditioned on $t$-th data set $D_t$, and $\textbf{I}_{t \times t}$ is an identity matrix of size $t$.

\subsection{Batch Bayesian Optimisation and Related Work}\label{sec:related}

BO is the task to find the global maximum of a blackbox function $f$:
\begin{align*}
    x^*_\text{true} = \mathop\mathrm{argmax}_{x \in \mathcal{X}} f(x),
\end{align*}
where $x^*_\text{true}$ represents the global optimum. BO is a model-based optimiser that typically uses a GP as a surrogate model \citep{osborne2009gaussian}. It uses GP predictive uncertainty to solve the blackbox optimisation problem, treating it as active learning to locate the global optimum. BO must balance the trade-off between exploitation (using current knowledge of the optimum from $m_t$) and exploration (exploring unseen optima considering uncertainty from $C_t$). Unnecessary exploration can lead to a slower convergence rate for the regret, defined as $r_t := f(x^*_\text{true}) - f(x_t)$, where $x_t$ is the $t$-th query point. The next query point is determined by maximising an acquisition function (AF), with the upper confidence bound (UCB; \citet{srinivas2009gaussian}) being a popular choice: ${\alpha_{f_t}}(x) := \mu_t(x) + \beta^{1/2}_t \sqrt{C_t(x,x)}$, where $\beta_t$ represents an optimisation hyperparameter. The rationale behind UCB is the decaying nature of the maximum information gain as more data is acquired (c.f., \citet{cover1999elements}). This decay is sublinear for popular kernels \citep{nemhauser1978analysis, krause2012near}, indicating progressively smaller changes for larger values of $t$, allowing us to demonstrate the \emph{no-regret property}, $\lim_{t \rightarrow \infty} \frac{r_t}{t} = 0$. Although there is a vast array of AFs, those with a proven no-regret property are limited to variants of either UCB, expected improvement (EI; \citet{bull2011convergence}; for the noiseless case), or TS, to the best of our knowledge.

However, most AFs are designed for sequential settings, and extending them to a batch setting often results in the loss of the no-regret property. Consequently, batch selection methods are mostly heuristic, yet they are widely accepted due to their practical significance and effectiveness. Batch BO methods can be categorised into the following two:

\paragraph{Optimisation-based approach.}
A prime example on this approach is the hallucination \citep{azimi2010batch}, which extends sequential methods by simulating the sequential process using a random sample from the GP predictive posterior. Despite its simplicity and empirical effectiveness, this method suffers from over-exploration due to mispecified GP models resulting from pseudo-labels (see Figure \ref{fig:demo}), and scalability issues with large batches. Each sequential query involves AF optimisation, essentially non-convex optimisation reliant on heuristic optimisers (e.g., CMA-ES; \citet{hansen2016cma}), thus introducing optimisation errors and overhead for each batch query. An alternative approach, using Monte Carlo (MC)-based AFs \citep{wilson2018maximizing, balandat2020botorch} for efficient parallel computations. However, as the authors noted, popular information-theoretic AFs \citep{hennig2015probabilistic, hernandez2014predictive, wang2017max} are not supported. Furthermore, the optimisation-based approach is challenged by a combinatorial explosion in scenarios involving discrete optimisation. As the number of categorical classes grows, the number of potential combinations becomes prohibitively large. Specifically, optimising for large batch sizes requires enumerating all conceivable permutations of \emph{both} batch samples and discrete variables, presenting a significant combinatorial challenge \citep{moss2021gibbon}. While recent work \citep{daulton2022bayesian} has tackled this, the proof is only applicable to sequential BO.

\paragraph{Thompson sampling-based approach.}
TS-based approaches can avoid the combinatorial and scalability issues through randomness. The AF of TS is $x_t = \mathop\mathrm{argmax}_{x \in \mathcal{X}} g(x)$, where $g \sim \mathcal{GP}(m_t, C_t)$ represents a function sample from the GP. This approach can be seen as sampling from the belief about the global optimum locations, $x_t \sim \mathbb{P}(\hat{x}^*_t \mid \mathcal{D}_t)$, where $\hat{x}^*_t$ is the estimated global optimum location. \citet{kandasamy2018parallelised} extended TS to batch BO, which still preserves the no-regret property. In batch BO, the key metric is the Bayesian regret (BR), defined by:
\begin{align*}
    \text{BR}(t) 
    &:= \mathbb{E}_{x_t \in \textbf{X}^n_t} [f(x^*_\text{true}) - f(x_t)]
\end{align*}
where $\textbf{X}^n_t$ is the batch TS samples drawn from $\mathbb{P}(\hat{x}^*_t \mid \mathcal{D}_t)$. By using the same rationale of UCB, when the maximum information gain is sublinear in the iteration $t$, the batch TS enjoys the no-regret properties for BR.

However, it faces limitations: incompatibility with other AFs and under-exploration. While the theory of batch TS depends on a well-specified GP, the common practice of using maximum likelihood estimation (MLE) for kernel hyperparameters does not ensure consistent estimation \citep{berkenkamp2019no, ziomek2024beyond}. This misspecification can invalidate no-regret property, leading to aggregated samples (see Figure \ref{fig:demo}), contrary to theoretical expectations. Although attempts have been made to address these issues \citep{nava2022diversified}, they often introduce significant overhead due to diversification, such as determinantal point process (DPP; \citet{kathuria2016batched}). Exact computation requires costly $\mathcal{O}(|\mathcal{X}| \cdot n^{6.5} + n^{9.5})$, and the best known inexact sampling with Markov chain Monte Carlo (MCMC) still demands $\mathcal{O}(n^5 \log n)$ MCMC steps \citep{rezaei2019a}.

Consequently, a versatile and lightweight batch BO algorithm remains elusive.

\section{Connection with Batch Uncertainty Sampling and Kernel Quadrature} \label{sec:kq}
In this section, we will demonstrate how KQ can provide a flexible and efficient solution for batch uncertainty sampling. We begin by introducing the concept of \emph{quantisation}, then establish the connection between batch uncertainty sampling and KQ.

\paragraph{Quantisation.}
Consider $\pi$ as a probability distribution defined over the domain $\mathcal{X}$. The task of \textit{quantisation} is to find a discrete distribution $\nu:= \frac{1}{n} \sum_{i=1}^n \delta_{x_i}$, which best approximates $\pi$ using $n$ representative points $x_i$. To tackle the quantisation task, one initially identifies an optimality criterion, typically based on a notion of \emph{discrepancy} between $\pi$ and $\nu$, and then devises an algorithm to approximately minimise this discrepancy. 

\paragraph{Kernel Quadrature.}
KQ is a numerical integration that computes the integral of a function $f$ within an RKHS $\mathcal{H}$ associated with a kernel $K$. Its goal is to approximate an, otherwise intractable, integral with a weighted sum. A KQ rule, $Q_{\pi, K}(n)$ is defined by weights $\textbf{w}^n = \{ w_i \}^n_{i=1}$ and points $\textbf{X}^n = \{ x_i \}^n_{i=1}$,
\begin{align}
Q_{\pi, K}(n):= \sum_{i=1}^n w_i f(x_i) \approx \int f(x) \text{d} \pi(x). \label{eq:kq}
\end{align}
The KQ rule can also be interpreted with a discrete distribution $\pi_\text{KQ}:= \sum_{i=1}^n w_i \delta_{x_i}$, namely, $Q_{\pi, K}(n) = \sum_{i=1}^n w_i f(x_i) = \int f(x) \text{d} \pi_\text{KQ}(x)$.
The \emph{worst-case error}, given $\pi$ and $\mathcal{H}$, is
\begin{align*}
\textup{wce}[Q_{\pi, K}(n)] := \sup_{\lVert f \rVert_{\mathcal{H}} \leq 1} \Bigg\lvert Q_{\pi, K}(n) - \int f(x) \text{d} \pi(x) \Bigg\rvert,
\end{align*}
and KQ aims to approximate $Q_{\pi, K}(n)$ that minimises this worst-case error,
\begin{align}
\textbf{X}^n, \textbf{w}^n &\approx \mathop\mathrm{argmin}_{\textbf{X}^n \subset \mathcal{X}, \textbf{w}^n \subset \mathbb{R}} \textup{wce} \left[ Q_{\pi, K}(n)\right]. \label{eq:kq-obj}
\end{align}
There are a vast list of KQ algorithms; herding \citep{chen2010super, bach2012on}, leverage score \citep{bach2017on}, DPP \citep{belhadji2019kernel}, continuous volume sampling \citep{belhadji2020a}, kernel thinning \citep{dwivedi2021kernel, dwivedi2022generalized}, to name a few.

\paragraph{Connection to Quantisation.}
The worst-case error can be considered a \textit{divergence} between $\pi$ and $\pi_\text{KQ}$. There is a theoretical link between KQ and quantisation, as KQ represents \textit{weighted} quantisation under the maximum mean discrepancy (MMD) metric \citep{karvonen2019kernel, teymur2021optimal}. MMD is a method for quantifying the divergence between two distributions \citep{sriperumbudur2010hilbert, muandet2017kernel}, defined as:
\begin{align*}
    \textup{MMD}_{\mathcal{H}}(\pi_\text{KQ}, \pi) := \Bigg\lVert \int K(\cdot, x)\text{d}\pi_\text{KQ}(x) - \int K(\cdot, x)\text{d}\pi(x) \Bigg\rVert_\mathcal{H},
\end{align*}
and we can rewrite as \citep{huszar2012optimally}:
\begin{align*}
    \textup{MMD}^2_\mathcal{H}(\pi_\text{KQ}, \pi) := \sup_{\lVert f \rVert_\mathcal{H} = 1} \Bigg\lvert \int f(x) \text{d} \pi_\text{KQ}(x) - \int f(x) \text{d} \pi(x) \Bigg\rvert^2.
\end{align*}
This squared formulation equates to the worst-case error: solving for KQ is the same to finding the discrete distribution $\pi_\text{KQ}$ that best approximates $\pi$ in terms of MMD. Note, KQ performs \emph{weighted} quantisation, differing from the previous unweighted quantisation.

\paragraph{Connection to Gaussian Process.}
Assuming a function $f_t$ is modelled by a GP, $f_t \sim \mathcal{GP}(m_t,C_t)$, with noisy observed points, $\mathcal{D}_t := (\boldsymbol{X}_t, \boldsymbol{Y}_t )$. Our objective is to estimate the expectation of the function $\hat{Z}:= \int f(x) \text{d} \pi(x)$. This scenario is referred to as Bayesian quadrature (BQ) \citep{o1991bayes, hennig2022probabilistic}, with integral estimates given by:
\begin{subequations}
\begin{align}
    \mathbb{E}_{f_t \sim \mathcal{GP}(m_t, C_t)} [ \hat{Z}] 
    &= \int m_t(x)\text{d} \pi(x)
    = \boldsymbol{z}^\top_t (\textbf{K}_{XX} + \sigma^2 \textbf{I}_{t \times t})^{-1} \textbf{Y}_t,
\label{eq:bq_mean}\\
    \mathbb{V}_{f_t \sim \mathcal{GP}(m_t, C_t)} [ \hat{Z} ]
    &= \int C_t(x, x^\prime) \text{d} \pi(x) \text{d} \pi(x^\prime)
    = z^\prime_t - \boldsymbol{z}^\top_t (\textbf{K}_{XX} + \sigma^2 \textbf{I}_{t \times t})^{-1} \boldsymbol{z}_t,
\label{eq:bq_var}
\end{align}
\end{subequations}
where $\boldsymbol{z}_t := \int K(x, \textbf{X}_t)\text{d} \pi(x)$ and $z^\prime_t := \int K(x, x^\prime) \text{d} \pi(x) \text{d} \pi(x^\prime)$ represent the kernel mean and variance, respectively. To enhance the accuracy of integration, it is desirable to minimise the uncertainty in the integral estimation as expressed in Eq.(\ref{eq:bq_var}). Therefore, Eq.(\ref{eq:bq_var}) can be regarded as the metric to assess the reduction in integral variance, which has been employed as the AF for BQ \citep{rasmussen2003bayesian, osborne2012active}. 

\paragraph{Connecting it All Together.}
\citet{huszar2012optimally} demonstrated that all of the worst-case error, MMD, and the integral variance are \emph{equivalent}. The BQ expectation in Eq.(\ref{eq:bq_mean}) is a weighted sum; $ \boldsymbol{z}^\top \boldsymbol{K}^{-1} \boldsymbol{y}_0 = \sum_{i=1}^n w_{BQ, i} y_i$, where $w_\text{BQ, j} := \sum_{i=1}^n \boldsymbol{z}^\top_{i} \boldsymbol{K}^{-1}_{i,j}$ and $\boldsymbol{K}^{-1} := (\textbf{K}_{XX} + \sigma^2 \textbf{I}_{t \times t})^{-1}$. 
These weights can be considered as forming a discrete distribution $\pi_\text{BQ} := \sum_{i=1}^n w_{\text{BQ}, i} \delta_{x_i}$, thereby allowing the integral variance estimation to be expressed as:
\begin{equation}
    \mathbb{V}_{f_t \sim \mathcal{GP}(m_t, C_t)} [ \hat{Z} ]
    = \textup{MMD}^2_\mathcal{H}(\pi_\text{BQ},\pi)
    = \inf_{\textbf{w}_\text{BQ}} \textup{wce}[Q_{\pi, C_t}]^2 \label{eq:kq-eq}
\end{equation}
for a fixed $X$,
where the kernel for MMD and KQ is the predictive covariance $C_t(\cdot, \cdot)$. This choice is due to $C_t(\cdot, \cdot)$ representing the posterior belief about $f$, which is expected to be more accurate than the prior belief represented by $K(\cdot, \cdot)$.

This demonstrates the close connection between KQ, GP, and quantisation. This equivalence shows that KQ is \emph{domain-aware batch uncertainty sampling}. Solving KQ is minimising the worst-case error, which is equivalent to minimising both MMD and the integral variance. MMD, being the quantisation, ensures that the resulting discrete points are spread over the distribution $\pi_t$ to approximate, representing domain-aware diversified sampling. The integral variance represents the expected uncertainty of GP, and its minimisation indicates batch uncertainty sampling. At first glance, minimising $\mathbb{V}_{f_t} [\hat{Z}]$ for uncertainty sampling might seem counterintuitive, as sequential uncertainty sampling typically maximises $C_t(x,x)$. However,  $\mathbb{V}_{f_t} [\hat{Z}]$ is a scalar value, not a function like $C_t(\cdot,\cdot)$, and computes a summary statistic indicating the quality of the selected nodes' approximation of the integral. Therefore, minimising this metric by selecting batch samples can be understood as batch uncertainty sampling. Importantly, KQ is the approximation of intractable integration, making it applicable to an arbitrary combination of $(K, \pi)$, unlike BQ\footnote{See Eq.(\ref{eq:kq}). While BQ needs the analytical kernel mean for the right hand side (Eqs.(\ref{eq:bq_mean})-(\ref{eq:bq_var})), KQ is approximating it with the weighted sum. Our previous work \citep{adachi2022fast} applied to batch BQ.}. 

\paragraph{In Summary.} 
A quantisation task can be regarded as a KQ task. The selected batch samples aim to minimise the divergence between the target distribution $\pi$ and the batch samples' distribution $\pi_{\text{KQ}}$. By employing the GP predictive covariance $C(\cdot, \cdot)$ as the kernel for the MMD, KQ transitions into batch exploration of GP uncertainty, concurrently minimising divergence from the target distribution. Consequently, batch construction through KQ offers a means to quantise the target distribution while incorporating uncertainty sampling.
The benefits of KQ are:
\begin{compactenum}
    \item Applicable to any kernel, given that the primary goal of the KQ objective is to approximate the intractable integral of the kernel function.
    \item Versatile across any domain, AFs, or constraints, provided the target distribution can be described as a probability measure $\pi$.
    \item To naturally produce diversified batch samples, and is able to assess its diversity using the widely recognised MMD criterion.
\end{compactenum}

\section{Batch Bayesian Optimisation as Quadrature}\label{sec:sober}
We now consider the application of KQ to the batch BO task. First, we demonstrate that the probabilistic lifting technique can transform the batch BO task into a KQ problem. Next, we explain how to solve this reinterpreted task using a KQ algorithm. Finally, we customise this general batch BO algorithm for varied cases.

\subsection{Probabilistic Lifting}
\begin{algorithm}
\caption{SOBER algorithm.}\label{alg:sober}
\begin{algorithmic}[1]
\REQUIRE domain prior $\pi_0$, initial data set $\mathcal{D}_0 = (\textbf{X}_0, \textbf{Y}_0)$, stopping criterion $\Delta_n$

\STATE $f_{t-1} \leftarrow \text{Initialise-GP}(\mathcal{D}_0)$\\%
\WHILE{$\mathbb{V}_{x}[\tilde{\pi}] < \Delta_n$}
\STATE $\pi_{t-1}, \alpha_{t-1}, C_{t-1} \leftarrow \text{Fit-GP-and-Update-}\pi(f_{t-1})$
\STATE $\textbf{X}^n_t, \textbf{w}^n_t, \mathbb{E}_{f_t}[\hat{Z}], \mathbb{V}_{x}[\tilde{\pi}] \leftarrow \text{KQ}(\pi_{t-1}, \alpha_{t-1}, C_{t-1})$
\STATE $\textbf{Y}^n_t = \text{Parallel-Query}(f_\text{oracle}(\textbf{X}^n_t))$
\STATE Update dataset $\mathcal{D}_t \leftarrow \mathcal{D}_{t-1} \cup (\textbf{X}^n_t, \textbf{Y}^n_t)$ and model $f_t \leftarrow \text{Update-GP}(f_{t-1}, \mathcal{D}_t)$.
\STATE Proceed next round $t \leftarrow t-1$.
\ENDWHILE
\RETURN global maximum estimate $\hat{y}^*_t = \max[\textbf{Y}_T]$, evidence estimate $\mathbb{E}_{f_t}[\hat{Z}]$
\end{algorithmic}
\end{algorithm}
To recast the batch BO task as a KQ task using probabilistic lifting, consider the dual formulation presented below:
\begin{equation}
    x^*_\text{true} \in \mathop\mathrm{argmax}_{x \in \mathcal{X}} f(x) \quad \xLeftrightarrow{\text{dual}} \quad
    \delta_{x^*_\text{true}} \in \mathop\mathrm{argmax}_{\pi \in \mathbb{P}(\mathcal{X})} \int f(x) \text{d}\pi(x),  \label{eq:dual}
\end{equation}
where $\delta_x$ denotes the delta distribution at $x$, making $\delta_{x^*_\text{true}}$ the point mass at the global maximum. Consequently, our goal aligns with the KQ objective in Eq.(\ref{eq:kq}), allowing the application of KQ algorithms to the batch BO task.

How do we interpret this dual formulation? We transform a non-convex optimisation problem, $\text{max} f(x)$, into an infinite-dimensional optimisation over the set of probability measures $\mathbb{P}(\mathcal{X})$. In other words, we do not consider pointwise updates: $\plim_{t \rightarrow \infty} x_t = x^*_\text{true}$ as in the conventional approach. Instead, we aim at \emph{distributional} updates, $\plim_{t \rightarrow \infty}\pi_t = \delta_{x^*_\text{true}}$, i.e., $\plim_{t \rightarrow \infty}\mathbb{E}_x[\pi_t] = x^*_\text{true}$, $\plim_{t \rightarrow \infty}\mathbb{V}_x[\pi_t] = 0$. This yields $\text{max} \int f(x) \text{d}\pi(x)$, making the non-convex objective $f$ linear and convex for $\pi$. This distributional formulation is attractive due to parallelisability and convexity, widely used in optimisation, from traditional primal-dual interior-point methods \citep{vandenberghe1996semidefinite, wright1997primal} to contemporary Bayesian machine learning theories \citep{rudi2020finding, wild2024rigorous}.

Algorithm \ref{alg:sober} outlines our algorithm, SOBER: Line 3 updates $\pi_{t-1}$ based on the GP $f_{t-1}$. Then, Line 4 employs the KQ algorithm to perform batch uncertainty sampling over $\pi_{t-1}$ to effectively reduce the uncertainty $C_{t-1}(\cdot, \cdot)$ by selecting batch points as quantisation $\pi_{KQ}$, where $\pi_\text{KQ} = \sum^n_{i=1} w_{KQ, i} \delta_{x_i}$, with $w_{KQ, i} \in \textbf{w}^n_t$ and $x_i \in \textbf{X}^n_t$. The resulting $\textbf{X}^n_t$ is the $n$-point batch BO samples, ensuring diversified batch uncertainty sampling over $\pi_{t-1}$. As illustrated in Figure \ref{fig:demo}, $\pi_{t-1}$ initially spans the domain, with $\textbf{X}^n_t$ diversified and progressively concentrating towards the global maximum over iterations. The variance $\mathbb{V}_x[\pi_t]$ becomes a natural choice of stopping criterion for a distributional convergence.

The next question is, \emph{what is $\pi$, and how do we update it?} Our probabilistic lifting transforms the original non-convex problem into an even more computationally demanding problem. Traditional algorithms often assume $f$ is polynomial, allowing for a further transition to moment space due to closed-form moments of $f$. However, with our black-box $f$ and the GP surrogate model's lack of closed-form kernel mean and variance for arbitrary $\pi$, a possible remedy is to presuppose a functional form for $\pi$. We propose two assumptions regarding $\pi$.

\paragraph{Thompson Sampling Interpretation (SOBER-TS).}
The first approach interprets $\pi$ as a probability distribution over the estimated global maxima $\hat{x}^*_t$, denoted as $\mathbb{P}(\hat{x}^*_t)$, where $\hat{x}^*_t$ represents the current estimation of the global maxima at $f_t$. The advantage of this perspective is that it aligns with existing theories on batch TS.

\begin{tcolorbox}[colback=white]
We unpack the interpretation of $\pi = \mathbb{P}(\hat{x}^*_t)$ step by step:
\begin{compactenum}
    \item $\hat{x}^*_t \sim \pi_t(x)$ is TS, namely $\hat{x}^*_t = \mathop\mathrm{argmax}_{x \in \mathcal{X}} g_t(x)$ and $g_t \sim \mathcal{GP}(m_t, C_t)$.
    \item $\pi_t$ is updated through conditioning $f_t$ with the new observations.
    \item The KQ approach selects batch samples that minimise the expected uncertainty $\mathbb{V}_{f_t}[\hat{Z}]$, allowing us to view $\textbf{X}^n_t$ as TS samples that most contribute to reducing uncertainty across the distribution $\mathbb{P}(\hat{x}^*_t)$.
\end{compactenum}
\end{tcolorbox}
How can we interpret applying KQ for batch TS with regard to the domain-aware batch uncertainty sampling? 
Firstly, the domain-aware means that the resulting KQ samples $\textbf{X}^n_t$ adhere to the original TS distribution, i.e., $\textbf{X}^n_t \sim \mathbb{P}(\hat{x}^*_t) = \pi_t(x)$. This method can thus be seen as a variant of batch TS, referenced in studies such as \citet{kandasamy2018parallelised, hernandez2017parallel, ren2024minimizing, dai2024batch}. 

Secondly, how does batch uncertainty sampling help the regret converge faster? As noted in §\ref{sec:related}, BR convergence rate depends on the spectral decay of maximum information gain defined as $\mathbb{I}(\textbf{Y}_t; f) = \mathbb{H}(\textbf{Y}_t) - \mathbb{H}(\textbf{Y}_t \mid f)$, quantifying the reduction in uncertainty about $f$ from revealing $\textbf{Y}_t$. For a GP, $\mathbb{I}(\textbf{Y}_t; f) = \mathbb{I}(\textbf{Y}_t; \textbf{f}_t) = \frac{1}{2} \log \lvert \textbf{I}_{t \times t} + \sigma^{-2} \textbf{K}_{XX}\rvert$, where $\textbf{f}_t := f(\textbf{X}_t)$. \citet{nemhauser1978analysis, krause2012near, srinivas2009gaussian} have shown that the information gain maximiser can be approximated by an uncertainty sampling with $(1-1/e)$ approximation guarantee. Thus, roughly speaking, the maximum information gain can be approximately seen as the largest predictive uncertainty $C_t$.
Ultimately, a faster spectral decay in the maximum information gain in iteration $t$ leads to faster BR convergence rate. This is the reason why typical BO theoretical paper has the kernel-specific convergence rate as each kernel has different spectral decay. Here, in later §\ref{sec:nystrom} and Theorem \ref{thm:plain}, we show that KQ batch uncertainty sampling can be understood as selecting the largest possible spectral decay of the given kernel $C_t$. Thus, roughly speaking, KQ is trying to select the samples with the largest possible information gain in batch $n$, thereby accelerating the spectral decay in iteration $t$. Moreover, in later Proposition \ref{prop:robustness}, we show that KQ is robust against model misspecification, thereby avoiding remaining stuck in local minima. Therefore, KQ can give robust, fast spectral decay sampling for batch TS. 

In distributional interpretation, KQ selects the points that minimise $\mathbb{V}_f[\hat{Z}]$, which minimises the predictive uncertainty $C_t$ over the current TS distribution $\pi_t$. The main source of variance $\mathbb{V}_x[\pi_{t-1}]$ is the predictive uncertainty $C_t$. Hence, the batch uncertainty sampling with KQ narrows the subsequent TS distribution $\pi_t$, steering it closer to $x^*_\text{true}$.

However, deriving the BR convergence rate of SOBER-TS is non-trivial because it requires an analysis of double spectral decay (one for batch $n$ in KQ, and one for iteration $t$ in maximum information gain). While the focus here is not on the SOBER-TS algorithm, future work may explore its BR convergence rate. Given the relationship between DPP and KQ \citep{belhadji2019kernel, belhadji2021analysis}, SOBER-TS is expected to match the convergence rate of DPP-TS \citep{nava2022diversified}, which has demonstrated a tighter Bayesian cumulative regret bound compared to standard batch TS approaches \citep{kandasamy2018parallelised}.

\paragraph{Likelihood-Free Inference Interpretation (SOBER-LFI).}
Figure \ref{fig:demo} illustrates that sampling directly from the TS distribution tends to remain stuck in local minima, contrary to theoretical expectations \citep{kandasamy2018parallelised}. This discrepancy arises from two primary causes: model misspecification and the non-closed-form nature of the distribution. Model misspecification leads to a mis-estimated distribution of $\hat{x}^*_t$. This causes sampling to be biased toward less promising regions, especially in the initial stages, as seen in Figure \ref{fig:demo}. This phenomenon is well-documented in the bandit literature \citep{simchowitz2021bayesian, kim2021doubly, aouali2023mixed}. Although exploratory adjustments through diversified sampling \citep{nava2022diversified} can alleviate this issue, they entail prohibitive computational costs. This is attributed to the challenge of sampling from low-probability regions due to the non-closed-form distribution, as random sampling $\hat{x}^*_t = \mathop\mathrm{argmax}_{x \in \mathcal{X}} g_t(x)$ is governed by its probability $\mathbb{P}(\hat{x}^*_t)$. Drawing samples from low-probability areas requires an exhaustive number of attempts (or luck). A closed-form expression enables more flexible sampling schemes, such as importance sampling.

To devise a more robust and fast sampling algorithm, we now consider a closed-form definition for $\pi$. Unlike previous bandit approaches that improve TS algorithms, we explore a non-TS approach. Given the uncertain nature of the global maximiser $\mathbb{P}(\hat{x}^*_t)$, $x^*_\text{true}$ could be at any location with values potentially exceeding the estimated maximum, denoted as $\hat{y}^*_t := \max f(\textbf{X}_t)$. With this insight, we can define $\pi_t(x)$ as follows:
\begin{align}
\pi_t(x) := \mathbb{P} \Big(f_t(x) \geq \hat{y}^*_t \mid \mathcal{D}_t \Big) \propto \Phi \left[\frac{m_t(x) - \hat{y}^*_t}{\sqrt{C_t(x,x)}} \right], \label{eq:lfi}
\end{align}
where $\Phi$ is the cumulative distribution function (CDF) of the standard normal distribution. This formulation aligns with the probability of improvement (PI; \citet{kushner1964new}), another widely-used AF in BO, offering a closed-form (albeit unnormalised) distribution that is easier to sample from than TS.

We interpret, for pedagogical reasons, the sequential update of $\pi$ as likelihood-free inference (LFI, \citet{hinton2002training, hyvarinen2009estimation, gutmann2011bregman, daolang2023learning})\footnote{LFI is often called `indirect inference' \citep{gourieroux1993indirect}, `synthetic likelihood' \citep{wood2010statistical, price2018bayesian}, or Approximate Bayesian Computation (ABC, \citet{csillery2010approximate, fujisawa2021gamma})}. LFI is particularly valuable when the analytical form of the likelihood is unavailable. In the context of BO, while we have a Gaussian likelihood $\mathbb{P}(y \mid x)$ for \emph{observed} data, the likelihood of the global optimum $\mathbb{P}(x^*_\text{true} \mid \hat{y}^*_t, \mathcal{D}_t)$ lacks an analytical form, because the true value of $x^*_\text{true}$ is unknown. This prevents the evaluation of the distance between a queried point $x_t$ and $x^*_\text{true}$. LFI addresses this by replacing the exact likelihood function with a `synthetic likelihood', which assesses divergence between observed and simulated data using summary statistics. This synthetic likelihood is updated with new observations and converges to the true likelihood as $t \rightarrow \infty$. This approach can be understood as a variant of the Bernstein-von Mises theorem \citep{van2000asymptotic}; under an infinite data scenario ($t \rightarrow \infty$), the Bayesian posterior converges to the MLE. The synthetic likelihood exhibits similar asymptotic behaviour \citep{pacchiardi2021generalized}. \citet{gutmann2016bayesian} demonstrated that PI function can be interpreted as a synthetic likelihood, where the CDF serves as summary statistic and $\hat{y}^*_t$ as the optimal threshold for LFI, asymptotically converging to standard Bayesian inference. \citet{wilson2024stopping} also rediscovered this (c.f., they framed Eq.(\ref{eq:lfi}) as $\epsilon$-optima, where $\epsilon = \hat{y}^*_t$). \citet{song2022a} extended on this LFI idea to allow non-GP surrogate models to employ for BO. Furthermore, \citet{wild2024rigorous} showed that the probabilistic lifting formulation could be understood within the framework of Bayesian inference. In this light we view $\pi_t(x)$ as the LFI synthetic likelihood of the global maximum $\mathbb{P}(\hat{x}^*_t \mid \hat{y}^*_t, \mathcal{D}_t)$, wherein $\pi_t(x)$ is updated and converging to $\plim_{t \rightarrow \infty}\mathbb{P}(\hat{x}^*_t \mid \hat{y}^*_t, \mathcal{D}_t) \rightarrow \mathbb{P}(x^*_\text{true}) = \delta_{x^*_\text{true}}$ as the iteration $t$ progresses, i.e., $\plim_{t \rightarrow \infty} \hat{y}^*_t \rightarrow y^*_\text{true} = f(x^*_\text{true})$. We denote this approach SOBER-LFI.
\begin{figure}
    \includegraphics[width=1\textwidth]{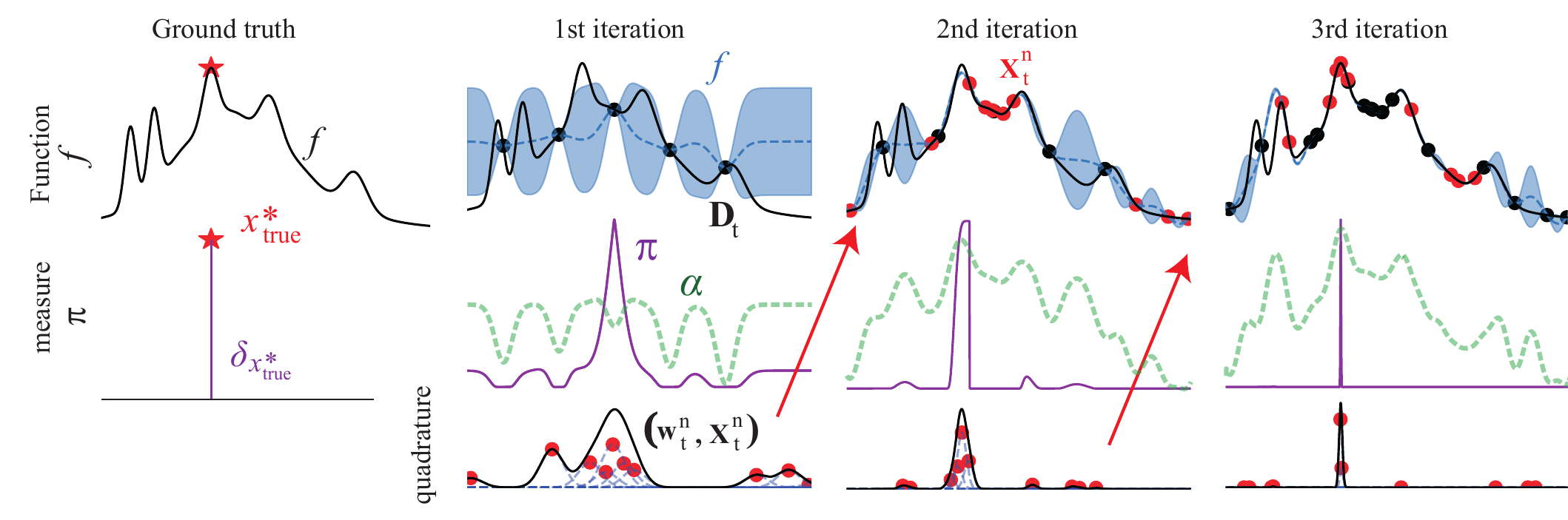}
    \caption{
    SOBER algorithm. Finding the location of global maximum $x^*_\text{true}$ is equivalent to finding the delta distribution $\delta_{x^*_\text{true}}$. Based on the surrogate $f_t$, we approximate the probability of global maximum $\mathbb{P}(\hat{x}^*_t)$ as $\pi$. We can also set the user-defined acquisition function $\alpha_t$ to adjust batch samples (UCB in this case).
    KQ algorithm gives a weighted point set $(\textbf{w}^n_t, \textbf{X}^n_t)$ that makes a discrete probability measure approximating $\pi$ (quantisation).
    Here, we have used a weighted kernel density estimation based on $(\textbf{w}^n_t, \textbf{X}^n_t)$ to approximately visualise the quantisation via KQ.
    Over iterations, $\pi$ shrinks toward global maximum, which ideally becomes the delta function in a single global maximum case.
    }
    \label{fig:sober}
    \vspace{-1em}
\end{figure}

Figure \ref{fig:sober} illustrates SOBER-LFI. The purple curve represents the synthetic likelihood $\pi_t$ as defined in Eq.(\ref{eq:lfi}), with the KQ algorithm selecting 10 batch samples that closely approximate the $\pi_t$ distribution and significantly contribute to reducing uncertainty. As the process progresses, the KQ-selected samples effectively reduce uncertainty: $\pi_{t=1}$ evolves into a much sharper $\pi_{t=2}$, centering around $x^*_\text{true}$, and by $\pi_{t=3}$, it nearly mirrors the delta distribution $\delta_{x^*_\text{true}}$. Remarkably, LFI ensures non-zero values in uncertain regions, enabling KQ to sample from these zones with very small probability in the last iteration, adhering to Cromwell's rule\footnote{According to this principle, the prior probability should remain non-zero for all possibilities. Since synthetic likelihood is updated sequentially, $\pi_{t-1}$ can be considered the prior for $\pi_t$.} \citep{jackman2009bayesian}. This behaviour is assured by the decaying, yet non-zero, nature of the CDF in Eq.(\ref{eq:lfi}), illustrating SOBER-LFI as a fusion of the exploitative PI $\pi_t$ and the explorative KQ batching algorithm.

Deriving the BR bound for this LFI framework presents more complexities compared to traditional TS strategies. To our knowledge, \citet{wang2018regret} stands alone in offering a simple regret analysis for PI. They predicated on the strong assumption that $y^*_\text{true} = f(x^*_\text{true})$ is known beforehand, and even if we accept it, the convergence rate does not hold a no-regret guarantee \citep{takeno2023posterior}. Nonetheless, although PI is not theoretically well-motivated, there are numerous successful studies in practice (e.g., \citet{bergstra2011algorithms, akiba2019optuna}).
Additionally, our focus diverges from mere maximisation, $x_t = \max_{x \in \mathcal{X}} \pi_t(x)$, to probabilistic sampling, $x_t \sim \pi_t(x)$. Hence, our SOBER-LFI aims diverge from those of sequential PI maximisation strategy. Interestingly, recent theoretical studies \citep{takeno2023posterior, ren2024minimizing} have established a tighter BR bound by integrating TS with PI, surpassing the results of conventional batch TS \citep{kandasamy2018parallelised}. Viewing TS as a randomness generator, merging TS with PI can be conceptually likened to PI supplemented by exploratory adjustments, aligning with our SOBER-LFI's concept. Thus, this emerging theoretical discourse might eventually elucidate the BR bounds for our SOBER-LFI algorithm. Still, despite their theoretical importance, we highlight that their methodologies are a variant of batch TS, thereby exhibiting limitations in misspecification robustness and universality, as depicted in Table \ref{tab:comparison}.

\subsection{Recombination: Kernel Quadrature Algorithm}
In this section, we reinterpret Eq.(\ref{eq:dual}) as a KQ problem and introduce a KQ algorithm to address it. Although any KQ method listed in §\ref{sec:kq} may be applied, we opt for the recombination approach \citep{hayakawa2022positively} to afford greater flexibility.

\subsubsection{Problem Setting of Kernel Quadrature}\label{sec:kq_conds}
For simplicity, we begin by considering the discrete optimisation scenario where $|\mathcal{X}| < \infty$. Suppose we have a kernel $C_t(\cdot, \cdot)$, which represents the GP posterior predictive covariance at the $t$-th iteration, and a set of $N$-point samples $\textbf{X}^N_t \in \mathcal{X}$, alternatively denoted by $x_i \in \textbf{X}^N_t$, associated with non-negative weights $\textbf{w}^N_t$, where $\{ w_i \in \textbf{w}^N_t \mid w_i > 0, \sum_{i=1}^N w_i = 1 \}$. We express this configuration as $\pi_t(x) := \sum_{i=1}^N w_i \delta_{x_i}$, treating it as a discrete distribution, or alternatively, as the ordered pair $(\textbf{w}^N_t, \textbf{X}^N_t)$.
The goal is to identify a weighted subset $\pi_\text{KQ}(x) := (\textbf{w}^n_t, \textbf{X}^n_t) = \sum_{j=1}^n w_j \delta_{x_j}$, where $n \ll N$, that minimises the MMD between $\pi_t$ and $\pi_\text{KQ}$, given the initial $\pi_t$ and the kernel $C_t(\cdot, \cdot)$. The quantised subset $\pi_\text{KQ}$, with $\textbf{X}^n_t$ being a subset of $\textbf{X}^N_t$, determines the batch samples for batch BO. In this discrete framework, we are equipped to compute the analytical worst-case error for an arbitrary kernel:
\begin{align*}
    \mathbb{E}_{f_{t-1}} [ \hat{Z}] &= \int m_{t-1}(x) \text{d} \pi_{t-1}(x) \approx {\textbf{w}^n_t}^\top m_{t-1}(\textbf{X}^n_t),\\
    \mathbb{V}_{f_{t-1}} [ \hat{Z} ]
    &= \textup{wce}[Q_{\pi_t, C_{t-1}}(n)],\\
    &= {\textbf{w}^n_t}^\top C_{t-1}(\textbf{X}^n_t, \textbf{X}^n_t) \textbf{w}^n_t - 2 {\textbf{w}^n_t}^\top C_{t-1}(\textbf{X}^n_t, \textbf{X}^N_t) \textbf{w}^N_t + {\textbf{w}^N_t}^\top C_{t-1}(\textbf{X}^N_t, \textbf{X}^N_t) \textbf{w}^N_t.
\end{align*}

Initially, $\pi_{t=0}$ represents the pool of unlabelled inputs, each assigned equal weights. As the iteration $t$ progresses, $\pi_t$ evolves into a subset of $\pi_{t=0}$, defined as $\textbf{X}^{N}_{t} = \textbf{X}^N_{t=0} \backslash \textbf{X}^{N}_{t-1}$\footnote{For brevity, we use $\textbf{X}^{N}_t$, despite $\textbf{X}^{N-tn}_t$ being more precise as $|\textbf{X}^{N}_{t}| = N - nt$. The same for $\textbf{w}^{N}_t$.}. The corresponding weights, $\textbf{w}^{N}_t$, are determined by whether $\pi$ is interpreted as TS or LFI.

\begin{tcolorbox}[colback=white]
Departing from the settings in previous works \citep{hayakawa2022positively,adachi2022fast}, we introduce the following conditions to our framework:
\begin{compactenum}
    \item[(a)] 
    A reward function $\alpha: \mathcal{X}\to\mathbb{R}$ is introduced to add flexibility, integrating additional considerations (e.g., AF). The objective is to maximise the expected reward ${\textbf{w}^n_t}^\top \alpha(\textbf{X}^n_t)$ while minimising the worst-case error $\textup{wce}[Q_{\pi_t, C_{t-1}}(n)]$.
\end{compactenum}
\end{tcolorbox}

\subsubsection{Kernel Quadrature via Nyström Approximation}\label{sec:nystrom}
While the Nyström method \citep{williams2000using, drineas2005on, kumar2012sampling} is commonly used for approximating large Gram matrices through low-rank matrices, it also offers a direct approach for approximating the kernel function itself. By selecting a set of $M$-points $X^M_t = \{ x_k \}_{k=1}^M \subset \mathcal{X}$, the Nyström approximation for $C_t(x, y)$ can be described as follows:
\begin{equation}
C_t(x, y) \approx \tilde{C}_t(x, y) := \sum_{j=1}^{n-1} \lambda_j^{-1}\varphi_j(x)\varphi_j(y),\label{eq:nys}
\end{equation}
where $\varphi_j(\cdot) := u^\top_j C_t(X_\text{nys}, \cdot)$, for $j=1,\ldots,n-1$, are termed \emph{test functions} and are derived from the broader $M$-dimensional space $\text{span}\{ C_t(x_k, \cdot)\}_{k=1}^M$.
To facilitate Eq.(\ref{eq:nys}), the optimal rank-$s$ approximation of the Gram matrix $C_t(\textbf{X}^M_t, \textbf{X}^M_t)= U \Lambda U^\top$ is determined via eigendecomposition. Here, $U = [u_1,\dotsc,u_M] \in \mathbb{R}^{M \times M}$ represents a real orthogonal matrix, and $\Lambda = \text{diag}(\lambda_1, \dotsc ,\lambda_M)$ consists of eigenvalues $\lambda_1 \geq \dotsc \geq \lambda_M \geq 0$. Eq.(\ref{eq:nys}) holds if $\lambda_s > 0$.

We can leverage these test functions for the integration estimator $\hat{Z} = \int f(x) \text{d} \pi_t(x)$. With a spectral decay in eigenvalues, the Nyström method efficiently approximates the original kernel function using a limited set of test functions. Defining $\boldsymbol\varphi = \{\varphi_1,\ldots,\varphi_{n-1} \}^\top$ as the vector of test functions spanning $\mathcal{H}_{\tilde{C}_t}$---the RKHS linked with the approximated kernel $\tilde{C}_t$---we assume knowledge of expectations, $\int \boldsymbol\varphi(x) \text{d} \pi_t(x) = {\textbf{w}^N_t}^\top \boldsymbol\varphi(\textbf{X}^N_t)$, is accessible.

This knowledge facilitates the construction of a convex quadrature:
\begin{align}
\sum_{j=1}^{n-1} w_j \varphi_i(x_j) = \int \boldsymbol\varphi(x) \text{d} \pi_t(x) \approx \int f(x) \text{d} \pi_t(x). \label{eq:testf}
\end{align}
Consequently, we can approximate the integral using $n-1$ test functions, interpreting Eq.(\ref{eq:testf}) as $n-1$ \emph{equality constraints} that both $w_j$ and $x_j$ must satisfy.

This method's advantage lies in its use of spectral decay information from the Gram matrix, promoting faster convergence. If the target function $f$ is smooth and exhibits fast spectral decay, then a small array of test functions can accurately represent the function, enhancing the efficiency of batch BO.

\subsubsection{Linear Programming Formulation}\label{sec:LP}
To solve the KQ task in Eq.(\ref{eq:kq-obj}), we introduce a linear programming (LP) problem. This problem is designed to simultaneously maximise the reward and minimise the worst-case error, modifying the approach from \citep{adachi2022fast}:
\begin{equation}
    \begin{aligned}
    \underset{\textbf{w} \subset  \mathbb{R}_{\geq 0}}{\text{max}} \quad
    &\textbf{w}^\top \alpha_t(\textbf{X}^N_t), \\
    \underset{\phantom{a}}{\text{subject to}} \quad 
    & (\textbf{w} - \textbf{w}^N_t)^\top \varphi_j(\textbf{X}^N_t) = \textbf{0},\\
    &\textbf{w}^\top \textbf{1} = 1, \ \ 
    \textbf{w} \geq \textbf{0},\ \ 
    |\textbf{w}|_0 = n, \forall j: 1 \le j \le n - 1,
    \end{aligned}\label{eq:lp}
\end{equation}
where $(\lambda_j, \varphi_j)$ are derived from the Nyström approximation (recall §\ref{sec:nystrom}), $\textbf{1} = [1,\ldots,1]^N$ signifies an all-ones vector, similarly for $\textbf{0}$, $\mathbb{R}_{\geq 0}$ represents the set of non-negative real numbers, and $|\cdot|_0$ indicates the count of non-zero entries.

\begin{tcolorbox}[colback=white]
The intuition of this formulation is as follows:
\begin{compactenum}
    \item[(1)] The solutions are defined by sparse weights $\textbf{w}$, where each non-zero weight corresponds to batch selection. The associated samples $\textbf{X}^N_t$ define the batch samples $\textbf{X}^n_t \subset \textbf{X}^N_t$. We denote the non-zero weights and their respective samples as the solution $\pi_\text{KQ} = (\textbf{w}^n_t, \textbf{X}^n_t)$, with the batch size being $|\textbf{X}^n_t| = |\textbf{w}|_0 = n$. Thus, this LP problem aims to subsample batch samples $\pi_\text{KQ}$ from the given discrete distribution $(\textbf{w}^N_t, \textbf{X}^N_t) \sim \pi_t$, effectively performing quantisation.
    \item[(2)] The goal is to maximise the expected reward $\alpha_t$. For example, if UCB is chosen as $\alpha_t$, it steers the batch samples towards the highest expected reward.
    \item[(3)] The first set of LP constraints\footnote{This is the worst-case error. $\bigl\lvert \textbf{w}^\top \boldsymbol\varphi(\textbf{X}^N_t) - {\textbf{w}^N_t}^\top \boldsymbol\varphi(\textbf{X}^N_t) \bigr\rvert \approx \bigl\lvert \int f(x) \text{d} \pi_\text{KQ}(x) - \int f(x) \text{d} \pi_t(x) \bigr\rvert$.} are equality constraints employing test functions from Eq.(\ref{eq:testf}). These $n-1$ equality constraints are stringent, significantly restricting the flexibility typically afforded by LP problems. Within this constrained space, the algorithm seeks the largest possible expected reward. When $\alpha(x) = 0$, the problem reverts to the standard KQ task, with the algorithm generating candidate sets $(\textbf{w}^n_t, \textbf{X}^n_t)$ that fulfill these constraints.
    \item[(4)] The other constraints ensure that the number of non-zero elements in $\textbf{w}$ matches the requested batch size $n$, maintaining convex and positive weights.
\end{compactenum}
\end{tcolorbox}
As such, solving the LP problem as defined in Eq.(\ref{eq:lp}) equates to addressing the KQ task using the Nyström approximation. This equivalence provides an efficient solution framework for the batch BO algorithm.

\subsubsection{Constrained Optimisation Formulation}\label{sec:LP_constrained}
We now show that a minor adjustment to SOBER can solve batch BO under unknown constraints. Consider global optimisation subject to unknown constraints, the problem setting proposed by \citet{gelbart2014bayesian}:
\begin{equation}
\begin{gathered}
    x^*_\text{true} = \mathop\mathrm{argmax}_{x \in \mathcal{X}} f(x),\\
    \text{subject to} \, \, g_l(x) \geq 0, \forall l \in [L],
\end{gathered}\label{eq:constraints}    
\end{equation}
where $f$ and each $g_l$ are unknown black-box functions\footnote{Eqs.(\ref{eq:constraints}) are only for the inequality constraints yet they can represent the equality constraint using two inequality constraints by bounding the upper and lower limit to be the same threshold.}. Notably, this approach \emph{permits constraint violations}, in contrast to settings that provide a known set of feasible solutions upfront to permit constraint breaches to be avoided completely \citep{sui2015safe}, an assumption invalidated by the black-box nature of our problem. In many practical scenarios, access to such feasible solutions is not available, nor is the feasibility of the problem itself. Additionally, stringent safety-critical constraints could trap these algorithms at local maxima. Nevertheless, our goal is to minimise the total violation incurred throughout the optimisation journey.

We model these black-box constraints using GPs, similarly to how we model the objective function. To maintain focus and brevity, we defer the comprehensive discussion on constraints modeling via GPs to \citet{gelbart2014bayesian, adachi2024adaptive}, concentrating instead on addressing Eqs.(\ref{eq:constraints}) with the given probabilistic models. Consequently, we assess the feasibility of constraints through a probabilistic lens, represented as $q_l$, rather than through a deterministic but unknown constraint function $g_l$.

\begin{tcolorbox}[colback=white]
Integrating the following conditions (b)(c)(d) with those in §\ref{sec:kq_conds}, we refine our tasks:
\begin{compactenum}
    \item[(b)] A tolerance for quadrature precision $\epsilon_\text{LP}$ is given. 
    \item[(c)] The specified batch size $n$ serves as an upper limit, with the actual batch size adaptively modified to achieve the desired precision within $\epsilon_\text{LP}$.
    \item[(d)] After selecting the batch querying points $(\textbf{w}^n_t, \textbf{X}^n_t)$, each point $x$ within $\textbf{X}^n_t$ is subject to the probabilistic constraint $q_l(x)$ (and violated w.p., $1-q_l(x)$). The functions $q_l:\mathcal{X}\to[0, 1]$ are modelled using GPs. Upon querying the true constraints $g_l(\textbf{X}^n_t)$, we isolate only those points that meet the constraints, along with their corresponding weights, denoted as $\tilde{\pi}_\text{KQ} = (\tilde{\textbf{w}}^n_t, \tilde{\mathbf{X}}^n_t)$. With $\tilde{\mathbf{X}}^n_t \subseteq \mathbf{X}^n_t$, this subset is used for both quadrature and batch BO\footnote{$\tilde{\mathbf{X}}^n_t = \mathbf{Z}^\top \mathbf{X}^n_t$, where $\mathbf{Z}$ is a vector of Bernoulli random variables with probabilities $q_l(\mathbf{X}^n_t)$}.
\end{compactenum}
\end{tcolorbox}

These conditions lead to a reformulated LP problem:
\begin{equation*}
\begin{aligned}
    \underset{\textbf{w} \subset \mathbb{R}_{\geq 0}}{\text{max}} \quad
    &\textbf{w}^\top \bigl[ \alpha_t(\textbf{X}^N_t) \odot \tilde{q}_t(\textbf{X}^N_t) \bigr], \\
    \underset{\phantom{a}}{\text{subject to}} \quad
    & \bigl\lvert(\textbf{w} - \textbf{w}^N_t)^\top \varphi_j(\textbf{X}^N_t) \bigr\rvert \leq \epsilon_\text{LP}\sqrt{\lambda_j / (n - 2) }
    \quad (1 \le {}^\forall j \le n - 2), \\
    & (\textbf{w} - \textbf{w}^N_t)^\top \tilde{q}_t(\textbf{X}^N_t) \ge 0,\ \ \textbf{w}^\top \textbf{1} = 1, \ \ 
    \textbf{w} \geq \textbf{0},\ \ 
    |\textbf{w}|_0 \le n,
\end{aligned}
\end{equation*}
where $\epsilon_\text{LP} \ge 0$ acts as a \emph{tolerance} level, representing the quadrature precision requirement---lower values indicate higher accuracy. The Hadamard product is denoted by $\odot$, and $\tilde{q}_t(\textbf{X}^N_t) = \bigodot_{l=1}^L q_l(\textbf{X}^N_t)$ signifies the joint probability of feasibility across all constraints $q_l$ at iteration $t$, with $\bigodot$ indicating multiple Hadamard products. In scenarios with a single constraint ($L = 1$), the joint feasibility mirrors the individual feasibility, making $\tilde{q}_t = q_1$.

\begin{tcolorbox}[colback=white]
The rationale behind these adjustments includes:
\begin{compactenum}
    \item[(1)] The objective is to maximise the product of the reward $\alpha_t$ and the joint feasibility $\tilde{q}_t$, guiding batch samples towards maximising expected `safe' reward.
    \item[(2)] We relaxed equality constraints to inequality ones to accommodate the $\epsilon_\text{LP}$ tolerance. The tolerance parameter $\epsilon_\text{LP}$ controls the trade-off between quadrature precision and the expansion of the solution space to find a larger objective.
    \item[(3)] Applicable for $n \geq 3$, this new LP formulation introduces an additional constraint, altering the Nyström approximation constraint count in Eq.(\ref{eq:testf}). It requires that the expected joint feasibility of the solution $\pi_\text{KQ}$ be equal to or greater than that of the initial candidate set $\pi_t$.
\end{compactenum}
\end{tcolorbox}
Consequently, the solution to this LP problem yields batch samples that not only comply with convex quadrature rules within the given tolerance but also maximise the expected safe reward. The balance between quadrature precision and reward optimisation is tunable via the single parameter $\epsilon_\text{LP}$, enabling the solution of constrained batch BO tasks within this framework. This approach parallels our prior work \citep{adachi2024adaptive} that focuses on the adaptive strategy.

\subsubsection{Adaptive Batch Sizes}\label{sec:adabatch}
Our focus now shifts to the aspect of adaptive batch sizes within our methodology. As illustrated in Table \ref{tab:comparison}, traditional algorithms maintain a constant batch size throughout experiments. This fixed strategy can be inefficient due to the dynamic balance between cost and speed---larger batches are more costly, smaller batches lead to slower wall-clock run-times---and the trade-off may change over the run (larger batches are often preferable earlier). To navigate this balance, we introduce an novel approach that adaptively adjusts batch sizes.

The concept is straightforward: we set a fixed tolerance for quadrature precision, $\epsilon_\text{LP}$, rather than fixing the batch size. This strategy allows batch sizes to automatically adjust to meet predetermined quadrature precision goals, similar to how standard optimisers cease operations based on convergence threshold. Our KQ strategy eliminates the need for exhaustive batch size trials across all possibilities. Furthermore, we extend this approach to constrained optimisation scenarios, treating constraint violations as decreases in precision requirements to subsequently adapt batch compositions.

\paragraph{No Constraints.}
The number of non-zero elements, $|\boldsymbol{w}|_0$, adjusts according to $\epsilon_\text{LP}$. The intuition of the batch size adaptivity is explicated as:
\begin{compactenum}
    \item Demanding higher precision decreases the quadrature error tolerance, necessitating a larger sample set for more accurate integration.
    \item Conversely, lower precision demands require fewer $|\boldsymbol{w}|_0$ to achieve the desired accuracy.
\end{compactenum}
Elaborating further, the batch size is tied to slack variables in LP solvers. 
An increase in $\epsilon_\text{LP}$ leads to the deactivation of some inequality constraints, as discussed by \citet{dantzig2002linear}. The active constraints determine the batch size, often resulting in sparse weights where $|\boldsymbol{w}|_0 < n$. Large fixed batch sizes become inefficient when fewer samples can fulfill the precision criteria. Thus, without resorting to brute-force searches across all potential batch sizes, we can identify the adaptive batch size $|\boldsymbol{w}|_0$.

$\epsilon_\text{LP}$ serves as a control lever for \emph{all} components: batch size, quadrature accuracy, and reward maximisation. Interestingly, its behaviour is not a monotonic decrease in its magnitude. As $\epsilon_\text{LP}$ approaches infinity, the batch size converges to 1, mirroring the sequential BO scenario. An increase in $\epsilon_\text{LP}$ reduces the batch size, as observed in §\ref{sec:adabatch_exp}, embodying a heuristic for adaptive batch sizes. While it ensures meeting a pre-defined worst-case error threshold, it does not promise the optimal outcome based on other established metrics like mutual information. However, as \citet{leskovec2007cost} notes, when greedily maximising mutual information under the weighted candidates and a budget constraint (limitation in the number of the total queries $T$), the approximation factor can be arbitrarily bad. Hence, even popular information-theoretic strategies also cannot achieve a solution within $1 - 1 / e$ of the optimal in our problem setting \citep{li2022batch}.

\paragraph{Under Constraints.}
\begin{figure}
  \centering
  \includegraphics[width=0.8\hsize]{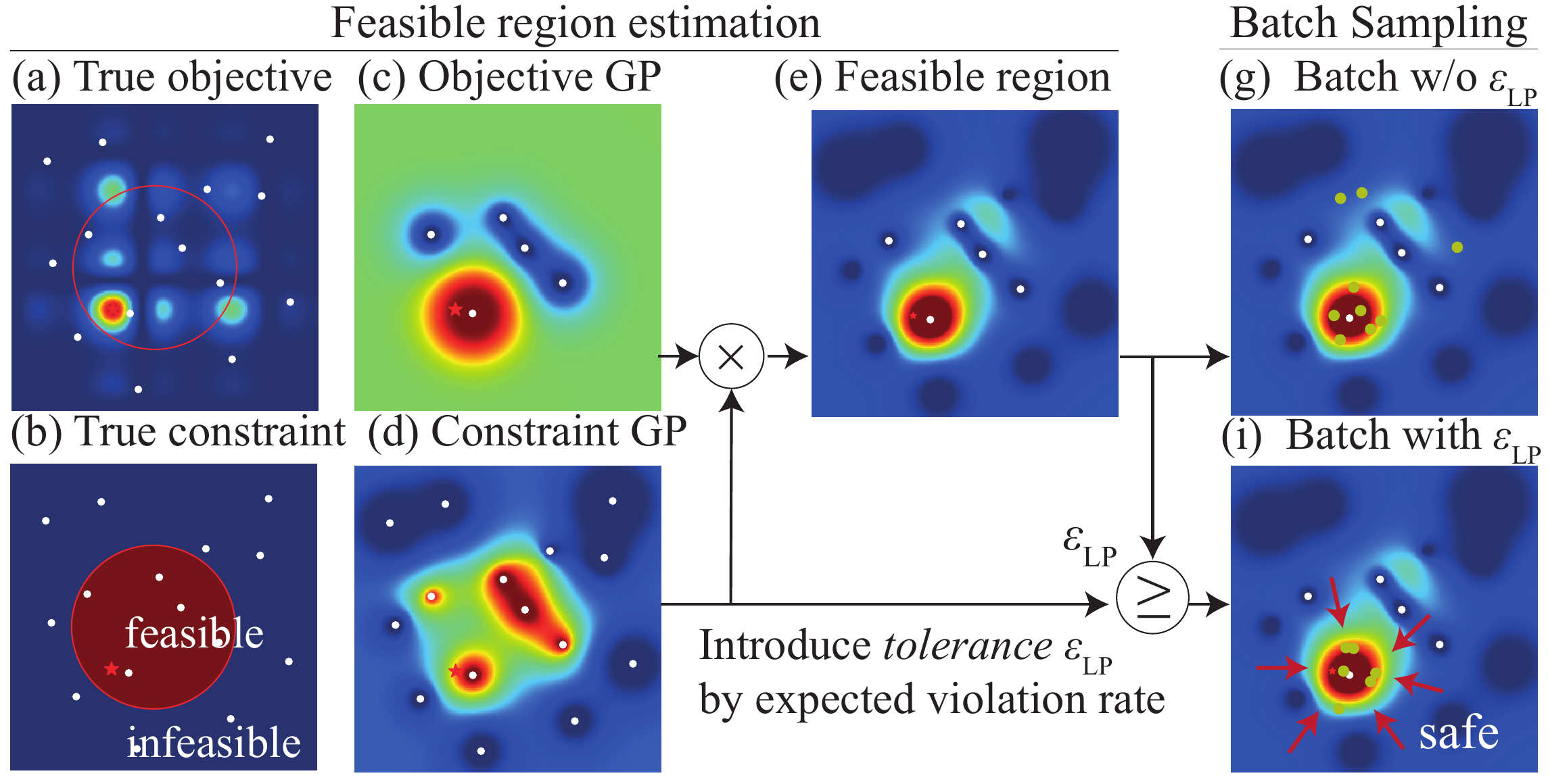}
  \caption{Constrained batch Bayesian optimisation. As the increased violation risk $\epsilon_\text{vio}$ propagates to the tolerance $\epsilon_\text{LP}$, reward maximisation is subsequently prioritised over quadrature, resulting in safe batch samples.}
  \label{fig:constrainedBO}
  \vspace{-1em}
\end{figure}
Adaptively adjusting batch size in the presence of probabilistic constraints $q_l$ is examined further. Given the uncertainty in accurately predicting the true constraint $g_l$, the candidate solution $\textbf{X}^N_t$ carries a violation risk. The \emph{expected} violation rate, $\epsilon_\text{vio}:= 1 - {\textbf{w}^N_t}^\top \tilde{q}_t(\textbf{X}^N_t)$, estimates the ratio of non-compliance. Infeasible points are excluded from the quadrature nodes for calculation, diminishing quadrature accuracy. $\epsilon_\text{vio}$, hence, represents an uncontrollable risk. High-risk scenarios demand cautious exploration to conserve valuable queries, suggesting smaller batches and selections where $\textbf{X}^N_t$ is likelier to meet the true constraint $g_l$. Low risk tolerates a more optimistic exploration approach.

We propose an \emph{adaptive} exploration strategy in response to varying risk levels, simply by setting $\epsilon_\text{LP} = \epsilon_\text{vio}$. This method permits automatic adjustment to exploration safety levels. With a high $\epsilon_\text{vio}$ indicating greater risk, a higher $\epsilon_\text{LP}$ leads to looser quadrature precision, smaller batch sizes, and a solution set more likely to comply with constraints\footnote{Lower $\epsilon_\text{LP}$ $\rightarrow$ looser LP inequality constraints $\rightarrow$ expanding solution space $\rightarrow$ larger LP objective $\rightarrow$ larger expected reward $\rightarrow$ larger feasibility $\rightarrow$ safer batch sampling, and vice versa.}. Thus, a higher $\epsilon_\text{LP}$ ensures safer batch sampling. Conversely, a lower risk level, indicated by a reduced $\epsilon_\text{vio}$, allows for setting a smaller $\epsilon_\text{LP}$, facilitating larger batches and more exploratory solutions. Figure \ref{fig:constrainedBO} showcases this adaptive mechanism: elevated risk $\epsilon_\text{vio}$ influences $\epsilon_\text{LP}$, leading to safer batch selections. This adaptive strategy effectively balances computational uncertainty and real-world risk, providing a flexible and automated means to navigate between ensuring safety and fostering exploration.

\subsection{Theoretical Bounds on Worst-case Error}
We now address the theoretical bounds on the worst-case errors in LP formulations, both with and without constraints (referenced in §\ref{sec:LP} and §\ref{sec:LP_constrained}).

\subsubsection{Kernel quadrature without constraints}\label{sec:emp_measure}
In the simplest scenario, we assess the worst-case error bounds within the context outlined in §\ref{sec:LP}. Here, rather than focusing on an \emph{exact} quadrature, we consider an \emph{approximate} quadrature using a MC estimate with a significantly large number of samples, denoted as the \emph{empirical measure}, $\tilde{\pi}_t := (\textbf{w}^N_t, \textbf{X}^N_t) \sim \pi_t$, where ${\textbf{w}^N_t}^\top \textbf{1} = 1$ and $\textbf{w}^N_t \geq 0$. The empirical measure represents a practical approximation of the true measure $\pi_t$, which could be a discrete distribution with an innumerable number of candidates $|\mathcal{X}|$, or a continuous distribution. This approach provides a versatile KQ method applicable across various samplable distributions, where the worst-case error is primarily influenced by the Nyström approximation error on the kernel and the distribution approximation error on the empirical measure. Studies such as those by \citet{drineas2005on,kumar2012sampling,hayakawa2023sampling} have thoroughly explored error bounds for this approximation:
\begin{theorem}\label{thm:plain}
    If an n-point convex quadrature $Q_{\pi_t, C_{t-1}}(n)$ satisfies $\pi_\text{KQ}(\varphi_j) = \tilde{\pi}_t(\varphi_j)$\footnote{For brevity, we denote $\pi(f) := \int f(x) \text{d}\pi(x)$.} for $1 \leq j \leq n - 1$ and $\pi_{KQ}\left(\sqrt{C_{t-1} - \tilde{C}_{t-1}}\right) \leq \tilde{\pi}_t\left(\sqrt{C_{t-1} - \tilde{C}_{t-1}}\right)$, then we have:
    \begin{equation*}
    \begin{aligned}
        \textup{wce}[Q_{\pi_t, C_{t-1}}(n)]
        &\leq \textup{MMD}_\mathcal{H}(\pi_\text{KQ}, \tilde{\pi}_t) + \textup{MMD}_\mathcal{H}(\tilde{\pi}_t, \pi_t),\\
        &\leq 2\underbrace{\tilde{\pi}_t\left(\sqrt{C_{t-1} - \tilde{C}_{t-1}}\right)}_\text{Nyström approximation} + \underbrace{\textup{MMD}_\mathcal{H}(\tilde{\pi}_t, \pi_t)}_\text{empirical measure},
    \end{aligned}
    \end{equation*}
\end{theorem}
which is taken from \citet[][Eq.~(13)]{hayakawa2023sampling}
Notably, if the finite number of candidates $|\mathcal{X}| = N$ is manageable, then $\tilde{\pi}_t = \pi_t$ and $\textup{MMD}_\mathcal{H}(\tilde{\pi}_t, \pi_t) = 0$, leaving the Nyström approximation error as the sole determinant of the error bound.

Crucially, this outcome is \emph{independent of dimensionality} but hinges on the kernel's spectral decay. Expanding the Nyström samples $M$ diminishes the first term, while increasing candidate numbers $N$ lessens the second term as it enhances $\tilde{\pi}_t$'s approximation of $\pi_t$. This indicates that enlarging $N$ and $M$ as permissible by time constraints can tighten the error bounds. However, the spectral decay's impact on maximum information gain in sequential BO is constrained by dimensionality, implying that the overall efficiency of batch-sequential algorithms is similarly affected by high-dimensional spaces, much like other BO methods.

\paragraph{Why Empirical Measure?}
The necessity for an empirical measure might seem superfluous when the exact integral $\mathbb{E}_{f_{t-1}}[\hat{Z}]$ is available for specific kernels, rendering the second term of the error negligible. There are several justifications for employing an empirical measure, and we now outline them.
\begin{tcolorbox}[colback=white]
The efficacy and rationale of using an empirical measure includes:
\begin{compactenum}
    \item[(1)] \textbf{Generality:} 
    An empirical measure can be formulated for any combination of $(\pi, K)$, extending beyond the scope of traditional BQ methods \citep{briol2019probabilistic}. This is particularly relevant as our $\pi_t$ undergoes sequential updates, potentially lacking a parametric form over $x$ (e.g., in TS scenarios).
    \item[(2)] \textbf{Hypercontractivity:} 
    Insights from the study of random convex hulls and hypercontractivity \citep{hayakawa2022hypercontractivity, hayakawa2023estimating} suggest that the requisite number of $N$ might be substantially lower than the actual search space, lending empirical support to the practicality of employing an empirical measure\footnote{Still, this is in a slightly different setting and has not been fully understood yet.}.
    \item[(3)] \textbf{Sequential $\pi$ update:} 
    With each iteration, $\pi$ narrows towards the global maximum, effectively reducing the number of viable candidates $N$ over time (as demonstrated in Figures \ref{fig:demo} and \ref{fig:sober}, where batch samples in later iterations aggregated around similar locations).
    \item[(4)] \textbf{Normalisation:} 
    The discrete nature of candidates simplifies normalisation, especially since our LFI $\pi_t$ in Eq.(\ref{eq:lfi}) is inherently unnormalised.
\end{compactenum}
\end{tcolorbox}

\subsubsection{Kernel quadrature with constraints}
Now consider the setting of constrained optimisation. Here, we introduce a tolerance for quadrature precision, necessitating an adjustment to the theoretical bound as follows:
\begin{prop}\label{prop:lp}
    Under the setting in the §\ref{sec:LP_constrained}, let 
    \rm$\textbf{w}_*$\it\ be the optimal solution of the LP, 
    and let \rm$\textbf{X}^n_t$\it\ be the subset of \rm$\textbf{X}^N_t$\it, 
    corresponding to the non-zero entries of \rm$\textbf{w}_*$\it\ 
    (denoted by \rm$\textbf{w}^n_t$\it).
    Suppose that \rm$\tilde{\textbf{X}}^n_t$\it\ 
    is given by a random subset of \rm$\textbf{X}^n_t$\it, where each point
    $x$ satisfies the constraints with probability $\tilde{q}_t(x)$,
    and let \rm$\tilde{\textbf{w}}^n_t$\it\ be the corresponding weights.
    Then, we have
    \rm\begin{equation}
        \mathbb{E}[
        {{\tilde{\textbf{w}}}^n_t}{}^\top
        \alpha_t(\tilde{\textbf{X}}^n_t)]
        \ge {\textbf{w}^n_t}^\top \bigl[ \alpha_t(\textbf{X}^n_t) \odot \tilde{q}_t(\textbf{X}^n_t)\bigr],
        \label{eq:lp-1}
    \end{equation}\it
    and, for any function $f_{t-1}$ in the RKHS with kernel $C_{t-1}$,
    \rm\begin{equation}
        \mathbb{E}\!\left[
        \left\lvert {\tilde{\textbf{w}}{}^n_t}{}^\top
        f_{t-1}(\tilde{\textbf{X}}^n_t) -
        {\textbf{w}^n_t}^\top f_{t-1}(\textbf{X}^n_t)\right\rvert\right] \le
        (\epsilon_\text{vio} K_{\max}
        + 2\epsilon_\text{nys} + \epsilon_\text{LP})\lVert f_{t-1} \rVert,
        \label{eq:lp-2}
    \end{equation}\it
    \it
    where $\lVert f_{t-1} \rVert$ is the RKHS norm of $f_{t-1}$,
    $K_{\max}:=\max_{x\in\textbf{X}^N_t} \sqrt{C_{t-1}(x, x)}$,
    and \rm$\epsilon_\text{vio}:= 1 - {\textbf{w}^N_t}^\top
    \tilde{q}_t(\textbf{X}^N_t)$\it\ is the expected violation rate
    with respect to the empirical measure given by
    \rm$(\textbf{w}^N_t, \textbf{X}^N_t)$\it, and $\epsilon_\text{nys}:=\max_{x\in\textbf{X}^N_t}
\lvert \tilde{C}_{t-1}(x, x) - C_{t-1}(x, x)\rvert^{1/2}$.
\end{prop}
The proof is given in Appendix \ref{app:proof}. This proposition elucidates that a quantitative approximation of the two tasks highlighted in §\ref{sec:LP_constrained} is achievable concurrently. It guarantees that, at minimum, the expected reward of the original batch is matched while ensuring the resulting measure $\tilde{\pi}_\text{KQ}$ (potentially non-probabilistic) conforms to the functions within the RKHS, all within a predefined error margin. This approach offers a quantitative framework for navigating the dual challenges of reward maximisation and constraint satisfaction in constrained optimisation scenarios.

\subsubsection{Robustness against misspecified RKHS}\label{sec:misspecification}
Finally, we address the robustness of our approach to misspecified RKHS. In BO, a common source of misspecification arises from inaccurate estimation in the hyperparameters of GPs. While the BO community has developed robust strategies \citep{berkenkamp2019no, bogunovic2021misspecified, ziomek2024beyond}, these are predominantly adaptations of the UCB and do not universally apply across all AFs. Conversely, the KQ community has thoroughly explored misspecification, offering robust estimations of worst-case errors for a broad range of conditions \citep{kanagawa2016convergence, oates2017control, karvonen2018bayes, kanagawa2020convergence}. Notably, the our KQ method also guarantees robustness against misspecification (Appendix B.4 in \citet{hayakawa2022positively}, using $\lvert\pi_\text{KQ}\rvert_\text{TV}=\lvert\tilde{\pi}_t\rvert_\text{TV} = 1$):
\begin{prop}\label{prop:robustness}
    Under the setting in the §\ref{sec:LP}, let $\mathcal{H}_{K_\text{mis}}$ be the misspecified RKHS and $\tilde{f} \in \mathcal{H}_{K_\text{mis}}$ be a function in the misspecified RKHS, and $\pi_\text{KQ}$ be a quadrature rule applied to a function $f \notin \mathcal{H}_{K_\text{mis}}$, leading only to the following bound via triangle equality and standard integral estimates:
    \begin{equation*}
    \begin{aligned}
        &\Bigg\lvert \int f(x) \text{d} \pi_\text{KQ}(x) - \int f(x) \text{d} \tilde{\pi}_t(x) \Bigg\rvert\\
        &\leq (\lvert\pi_\text{KQ}\rvert_\text{TV}+\lvert\tilde{\pi}_t\rvert_\text{TV})\sup_{x \in \mathcal{X}} \lvert f(x) - \tilde{f}(x) \rvert  + \lVert \tilde{f} \rVert_{\mathcal{H}_{K_\text{mis}}} \textup{wce}[Q_{\tilde{\pi}_t, K_\text{mis}}(n)]\\
        &=2\sup_{x \in \mathcal{X}} \lvert f(x) - \tilde{f}(x) \rvert  + \lVert \tilde{f} \rVert_{\mathcal{H}_{K_\text{mis}}} \textup{wce}[Q_{\tilde{\pi}_t, K_\text{mis}}(n)].
    \end{aligned}
    \end{equation*} 
\end{prop}

The first inequality in Proposition \ref{prop:robustness} highlights the advantage of employing convex weights within the KQ rule. Non-convex weights can inflate the total variation $\lvert \pi_\text{KQ} \rvert_\textup{TV}$, potentially resulting in significant integration errors. Unlike traditional BQ, which adopts negative weights and thus suffer from misspecification challenges \citep{huszar2012optimally}, the use of convex weights as delineated here mitigates such risks at least within uniform bounds, underscoring the robustness of our KQ approach against RKHS misspecification.

\subsection{In Practice. How to Solve LP and Apply SOBER}
\subsubsection{How to solve LP problem}\label{sec:inpractice}
We introduce the following two algorithms to solve the above LP problems. We detail two algorithms for addressing LP problems: the \emph{recombination} algorithm for unconstrained settings (§\ref{sec:LP}) and the \emph{LP solver} for constrained scenarios (§\ref{sec:LP_constrained}).

\paragraph{Recombination.}
The recombination algorithm \citep{litterer2012high, tch15, cosentino2020randomized, hayakawa2022positively} offers an efficient algorithm to solve scenarios that meet the following conditions: (i) absence of constraints, (ii) exact solution requirements ($\epsilon_\text{LP} = 0$), and (iii) a fixed batch size ($|\textbf{X}^n_t| = n$). Recombination is an efficient solver for special linear programming task, differing from a general solver like the simplex method. Recombination leverages Carathéodory's theorem for fast computation through mere matrix operations, with further details available in \citet{tch15}. Its computational complexity, $\mathcal{O}(C_\varphi N + n^3 \log(N/n))$, where $C_\varphi$ represents the cost of evaluating $(\varphi_j)^{n-1}_{j=1}$ at any given point, is the most efficient for the stated conditions. Given that typical batch BO settings align with these prerequisites, recombination is the recommended primary solver.

\paragraph{General LP solver.}
For broader applications, including those with constraints, a general LP solver using the simplex method becomes relevant. In unconstrained scenarios with adaptive batch sizes, setting $\epsilon_\text{LP}$ to a minimal value like $10^{-8}$ is recommended to minimise numerical errors linked to floating-point precision limits. In the presence of constraints, $\epsilon_\text{LP}$ auto-adjusts based on the estimated risk level, $\epsilon_\text{LP} = \epsilon_\text{vio}$, eliminating the need for manual tolerance settings. Additionally, the incorporation of randomised singular value decomposition (SVD; \citet{hal11}) for Nyström approximation enhances computational speed, with practical performance surpassing the theoretical time complexity of $\mathcal{O}(N M + M^2 \log n + M n^2 \log(N / n))$ as noted in \citet{hayakawa2022positively}. This approach ensures that LP problems, whether constrained or not, can be solved with efficiency and precision, making it an essential component of the SOBER algorithm's practical application.

\subsubsection{How to sample from \texorpdfstring{$\pi_t$}{p}}
\begin{algorithm}
\caption{Sequential importance resampling.}\label{alg:pi_sampling}
\begin{algorithmic}[1]
\REQUIRE domain prior $\pi_0$, target distribution $\pi_t$

\STATE Initial sampling $\tilde{\textbf{X}}^N_t \sim \pi_0$:
\STATE Compute normalised importance weights $\tilde{\textbf{w}}^N_t = \textup{Normalise}(\pi_t(\tilde{\textbf{X}}^N_t) / \pi_0(\tilde{\textbf{X}}^N_t))$
\STATE Maximum likelihood Estimate $\tilde{\pi}_0 \leftarrow \textup{MLE}(\pi_0, \tilde{\textbf{X}}^N_t, \tilde{\textbf{w}}^N_t)$
\STATE Refined resampling $\textbf{X}^N_t \sim \tilde{\pi}_0$:
\STATE Compute normalised importance weights $\textbf{w}^N_t = \textup{Normalise}(\pi_t(\textbf{X}^N_t) / \tilde{\pi}_0(\textbf{X}^N_t))$
\RETURN empirical measure $\tilde{\pi}_t = (\textbf{w}^N_t, \textbf{X}^N_t)$
\end{algorithmic}
\end{algorithm}
As elucidated in §\ref{sec:emp_measure}, we use the empirical measure $\tilde{\pi}_t(x)$ to approximate $\pi_t$. Essentially, constructing empirical measure $\tilde{\pi}_t(x) = (\textbf{w}^N_t, \textbf{X}^N_t) \sim \pi_t(x)$ is sampling from $\pi_t$. For directly samplable distributions $\pi_t(x)$, we generate i.i.d. samples from $\pi_t(x)$ and assign $\textbf{w}^N_t$ as a discrete uniform distribution $\{w_i \in \textbf{w}^N_t \mid w_i = 1 / N, \forall i \in [N] \}$. In cases where direct sampling is not feasible, the classic \emph{sequential importance resampling (SIR)} method \citep{kitagawa1993monte} becomes invaluable, as detailed in subsequent sections and Algorithm \ref{alg:pi_sampling}. Our Python library simplifies this process, allowing users to easily select the domain corresponding to their optimisation objectives without diving into the technicalities.

\paragraph{Thompson Sampling.}
The TS variant of SOBER, primarily for baseline comparison, incurs a time-intensive sampling process, a challenge shared by other TS-based algorithms \citep{nava2022diversified}. Constructing the empirical measure $\tilde{\pi}_t$ with TS is straightforward: draw $N$ exhaustive TS samples and set $\textbf{w}^N_t$ as a discrete uniform distribution. Although recent advancements in fast posterior sampling methods \citep{wilson2020efficiently, lin2024sampling} slightly mitigate the exhaustive nature of this sampling process, the SOBER-LFI procedures are markedly simpler and more efficient.

\paragraph{Discrete Domain.}
Generally, a discrete domain implies a discrete distribution with equal weights, $\nu = 1/N \sum_{i=1}^N \delta_{x_i}$. If all candidates $\mathcal{X}$ can be enumerated (e.g., in drug discovery, where all viable candidates are known), using the parametric distribution $\mathbb{P}(\mathcal{X}) = \mathcal{U}(\mathcal{X})$, where $\mathcal{U}$ represents the discrete uniform distribution, maintains generality without loss of precision. Herein, the distribution class generating $x \sim \mathbb{P}(\mathcal{X})$ is termed \emph{the domain prior} $\pi_0(x) = \mathbb{P}(\mathcal{X})$. In combinatorial optimisation of binary variables, for instance, the domain prior is essentially a Bernoulli distribution with equal weights. For discrete variables with categories (e.g., choices among $\{ 0,1,2,3,4 \}$), this equates to a categorical distribution with equal weights, enabling the use of Bernoulli, categorical, or general discrete distributions as the domain prior $\pi_0$ through SIR to construct $\tilde{\pi}_t$.

\paragraph{Continuous Domain.}
The continuous domain presents more complex challenges for SOBER-LFI due to the absence of a defined parametric form for the domain prior. For computational simplicity and efficiency, we opt for a Gaussian Mixture Model (GMM) \citep{xuan2001algorithms} as the samplable parametric model for the domain prior $\pi_0$. Given the universal approximation capabilities of GMM \citep{stergiopoulos2017advanced}, it serves as a suitable proposal distribution for the SIR procedure. Note that our end goal is importance sampling $\tilde{\pi}_t$, negating the need for an exact match of GMM proposal distribution to $\pi_t$.

\paragraph{Mixed Domain.}
In scenarios with a mixture of continuous and discrete variables, assuming independent distributions for each segment is practical. For example, if the initial three dimensions are continuous and the subsequent two are binary, the domain prior could be modelled as a product of continuous and discrete distributions $\pi_0 \propto \textup{GMM}(\textbf{X}_{:3}) \cdot \textup{B}(\textbf{X}_{4:5})$, where $\textup{GMM}$ and $\textup{B}$ represent the GMM and Bernoulli distribution, respectively, and $\textbf{X}_{k:k'}$ specifies the input dimensions from $k$ to $k'$.

\paragraph{Expert Knowledge.}
Engagement with human experts can yield beliefs about the global maximum's location $\mathbb{P}(\hat{x}^*_t)$, derived from experience, expertise, or prior knowledge (e.g., \citet{hvarfner2022pi, adachi2024looping}). In such instances, this expert model is directly employed as the domain prior $\pi_0$, facilitating a tailored approach to leveraging domain-specific insights in optimisation tasks.

\paragraph{Batch Active Learning and Bayesian Quadrature Tasks.}
In the context of pool-based batch active learning (AL), the primary goal is the information gain of the model parameters for more accurate prediction. Specifically, within a GP model, this often translates to batch uncertainty sampling. We usually operate under the assumption that a set of unlabelled candidates, $\mathcal{X}$, is provided and can be fully enumerated. Consequently, in the batch AL scenario, our target distribution, $\pi_t$, is defined as a uniform distribution over the available candidates, $\pi_t := \mathcal{U}(\mathcal{X})$. Conversely, batch BQ presupposes a prior distribution, $\pi_0$, that remains constant throughout the process. Therefore, in the batch BQ framework, our distribution $\pi_t$ aligns with this predefined prior, $\pi_t = \pi_0$. In both tasks, the distribution $\pi_t$ is stationary. As such, the main distinction between batch AL, BQ, and BO lies in mere definition of $\pi_t$, and our SOBER framework is applicable for these tasks.

\section{Experiments}\label{sec:experiments}
\begin{table}
    \centering
    \resizebox{1\textwidth}{!}{
    \begin{tabular}{lccccccccccc}
    \toprule
    & & & & \multicolumn{2}{c}{dimensions $d$}
    & & & & & \multicolumn{2}{c}{domain prior $\pi_0$} \\
    \cmidrule(lr){5-6} \cmidrule(lr){11-12}
    experiments &
    task & \begin{tabular}[c]{@{}c@{}}syn.\\ /real\end{tabular} & objective & cont. & disc. & \begin{tabular}[c]{@{}c@{}}space\\ $\mathcal{X}$\end{tabular} & 
    \begin{tabular}[c]{@{}c@{}}batch\\ $n$\end{tabular} &
    \begin{tabular}[c]{@{}c@{}}const-\\ raint $L$ \end{tabular} &
    \begin{tabular}[c]{@{}c@{}}kernel\\ $K$\end{tabular} & cont. & disc. \\
    \midrule
    Branin-Hoo & BO & syn. & $\max(f)$ & 2 & - & cont. & $=30$ & - & RBF & $\mathcal{U}(-3,2)$ & - \\
    Ackley & BO & syn. & $\log_{10} \min(f)$ & 3 & 20 & mixed & $=200$ & - & RBF & $\mathcal{U}(-1,1)$ & Bernoul. \\
    Rosenbrock & BO & syn.& $\log_{10} \min(f)$
    & 1 & 6 & mixed & $=100$ &- &  RBF & $\mathcal{U}(-4,11)$ & Categor. \\
    Hartmann & BO & syn.   & $\log_{10} \max(f)$ & 6 & - & cont. & $=100$ & - &RBF & $\mathcal{U}(0,1)$ & - \\
    Snekel      &  BO & syn.    & $\log_{10} \max(f)$ & 4 & - & cont. & $=100$ & - & RBF & $\mathcal{U}(0,10)$ & - \\
    \cmidrule(lr){3-12}
    Pest       &    BO & real. & $\min(f)$ & - & 15 & disc. & $=200$ & - & RBF & - & Categor. \\
    MaxSat      &   BO & real. & $\min(f)$                     & - & 28 & disc. & $=200$ & - & RBF & - & Bernoul. \\
    Ising       &   BO & real.  & $\log_{10} \min(f)$                     & - & 24 & disc. & $=100$ & - & RBF & - & Bernoul. \\
    SVM         &   BO & real. & $\min(f)$                     & 3 & 20 & mixed & $=200$ & - & RBF & $\mathcal{U}(0,1)$ & Bernoul. \\
    Malaria     &   BO & real.  & $ \log_{10} \min(f)$                     & \multicolumn{2}{c}{molecule} & disc. & $=100$ & - & Tanimoto & - & Categor.\\
    Solvent     &   BO & real. & $- \max\log_{10}f$                     & \multicolumn{2}{c}{molecule} & disc. & $=200$ & - & Tanimoto & - & Categor. \\
    \cmidrule(lr){2-12}
    Branin-Hoo & cBO & syn. & $\max(f)$ & 2 & - & cont. & $\leq 20$ & 2 & RBF & $\mathcal{U}(-3,2)$ & - \\
    Ackley & cBO & syn. & $\log_{10} \min(f)$ & 3 & 20 & mixed & $\leq 200$ & 2 & RBF & $\mathcal{U}(-1,1)$ & Bernoul. \\
    Hartmann & cBO & syn. & $\log_{10} \textup{regret}$ & 6 & - & cont. & $\leq 5$ & 2 &RBF & $\mathcal{U}(0,1)$ & - \\
    Pest       &    cBO & real. & $\min(f)$ & - & 15 & disc. & $\leq 200$ & 2 & RBF & - & Categor. \\
    Malaria     &   cBO & real.  & $ \log_{10} \min(f)$                     & \multicolumn{2}{c}{molecule} & disc. & $\leq 100$ & 4 & Tanimoto & - & Categor.\\
    FindFixer   &   cBO & real.  & $ \max(f)$                     & \multicolumn{2}{c}{node} & graph & $\leq 100$ & 3 & graph & - & Categor.\\
    TeamOpt   &  cBO & real.  & $ \log_{10} \textup{regret}$                     & \multicolumn{2}{c}{subgraph} & graph & $\leq 100$ & 3 & graph & - & Categor.\\
    \cmidrule(lr){2-12}
    2 RC        &   BQ & real.  & $\int f(x) \text{d} \pi_0(x)$ & 6 & - & cont. & $=100$ & - & RBF & Gaussian & - \\
    5 RC        &   BQ & real.  & $\int f(x) \text{d} \pi_0(x)$     & 12 & - & cont. & $=100$ & - & RBF & Gaussian & -\\
    \bottomrule
    \end{tabular}
    }
    \caption{Experimental Setup. Task: either BO, constrained BO (cBO), or BQ. Syn./real: synthetic or real-world. Dimensions:  the number of dimensions over input space categorised into continuous (cont.), discrete (disc.). Batch: the fixed batch size $=n$ or upper bound of adaptive batch size $\leq n$. Constraint: the number of constraints $L$. Prior: the domain prior $\pi_0$, Bernoul. and Categor.: Bernoulli and categorical distributions with equal weights. Special kernels are used: Tanimoto kernel \citep{ralaivola2005graph} for molecules and the diffusion graph kernel \citep{zhi2023gaussian} for graphs.
    }
    \label{tab:exp}
    \vspace{-1em}
\end{table}
We now move into the evaluation of our algorithm, SOBER, through both synthetic and real-world examples. Initially, we empirically analyse our proposed methodologies, focusing on measure convergence, robustness against misspecified RKHS, scalability, hyperparameter sensitivity, and adaptability of batch sizes. Subsequently, we juxtapose SOBER's performance in terms of regret convergence with that of popular baselines.

Our assessment spans \emph{20 experiments, benchmarked against 17 baselines}; 14 baselines for BO;
random, 
batch TS \citep{kandasamy2015high}, 
decoupled TS \citep{wilson2020efficiently}, 
DPP-TS \citep{nava2022diversified}, 
TurBO \citep{eriksson2019scalable}, 
GIBBON \citep{moss2021gibbon}, 
hallucination \citep{azimi2010batch}, 
local penalisation (LP\footnote{Only within the exerimental section, LP refers to local penalisation, and we use LP for linear programming for other sections.}; \citet{gonzalez2016batch}), 
B3O \citep{nguyen2016budgeted},
cEI \citep{benjamin2019constraned}, 
PESC \citep{hernandez2016a}, 
SCBO \citep{eriksson2021scalable}, 
cTS \citep{eriksson2021scalable}, 
and PropertyDAG \citep{park2022propertydag},
and 3 baselines for BQ; 
batchWSABI \citep{wagstaff2018batch}, 
BASQ \citep{adachi2022fast}, and 
logBASQ \citep{adachi2023bayesian}. 

The 20 experimental setups are detailed in Table \ref{tab:exp}, comprising 11 BO experiments (5 synthetic and 6 real-world data sets), 7 constrained BO experiments (3 synthetic and 4 real-world data sets), and 2 real-world experiments tailored to BQ. Comprehensive experimental methodologies are provided in Appendix \ref{app:exp}. Our implementations leverage PyTorch-based libraries \citep{paszke2019pytorch, gardner2018gpytorch, balandat2020botorch, griffiths2024gauche}, with all tests averaged over 10 iterations and executed in parallel on multicore CPUs for fair comparison. We note that GPU could further enhance SOBER's performance. Experimental outcomes are presented as the mean $\pm$ standard error of the mean, adhering to default SOBER hyperparameters $N=20,000, M=500$, unless otherwise specified.

To facilitate comparisons in discrete or mixed domains where certain algorithms (e.g., TurBO, GIBBON, Hallucination, and LP) encounter challenges due to combinatorial complexities, we employ a thresholding approach, optimising discrete variables as continuous ones and then categorising solutions through nearest neighbours. For the special yet popular tasks, such as drug discovery and graph tasks, non-Euclidean spaces or specialised kernels preclude the application of most algorithms.

\subsection{Measure Convergence Analysis}
\begin{figure}
  \centering
  \includegraphics[width=0.7\textwidth]{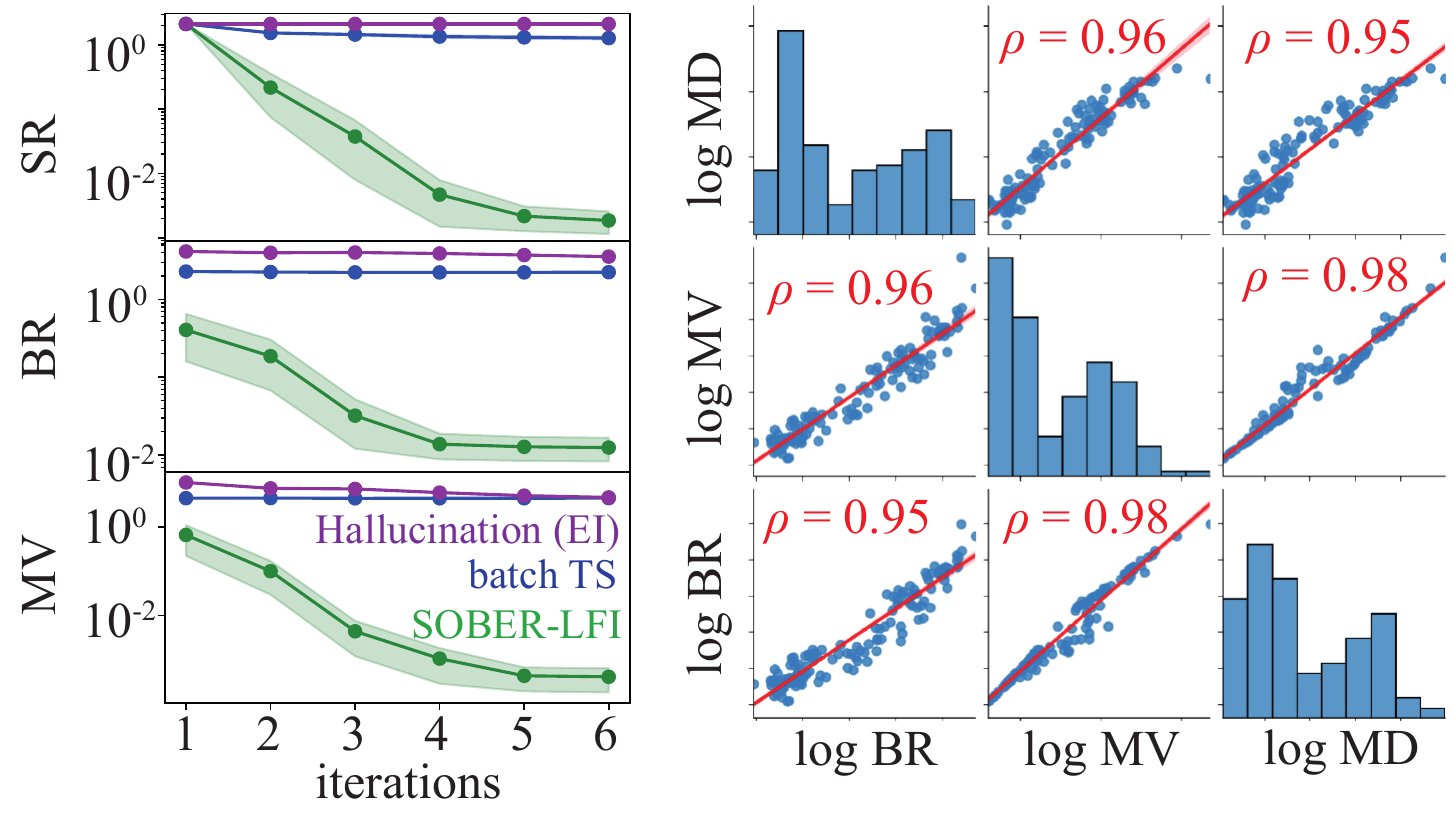}
  \caption{Correlations between Bayesian regret (BR) and measure optimisation. (Left) the convergence of simple regret (SR), BR, and mean variance (MV) for three batching methods. (Right) the linear correlations between mean distance (MD), MV, and BR.}
  \label{fig:corr}
  \vspace{-1em}
\end{figure}

We empirically investigated the relationship between regret and $\pi_t$ convergence. Recall that the empirical measure $\tilde{\pi}_t = (\textbf{w}^N_t, \textbf{X}^N_t)$ is sampled from $\pi_t$, and the KQ rule $\pi_\text{KQ} = (\textbf{w}^n_t, \textbf{X}^n_t)$ is the subset further extracted from $\tilde{\pi}_t$. As such, all measures approximate the same distribution, $\pi_t \sim \tilde{\pi}_t \sim \pi_\text{KQ}$, with only the level of discretisation differing. We consider the following two metrics for $\pi_t$ convergence: mean Euclidean distance (MD) $|x^*_\text{true} - \mathbb{E}_x[\pi_t(x)]|$ and mean variance (MV) $\mathbb{V}_x[\pi_t(x)]$, which can be approximated using $\tilde{\pi}_t(x)$:
\begin{align*}
    \mathbb{V}_x[\pi_t(x)]
        &\approx {\textbf{w}^N_t}^\top \text{diag} \Big[ (\textbf{X}^N_t - \mathbb{E}_x[\tilde{\pi}_t(x)])^\top (\textbf{X}^N_t - \mathbb{E}[\tilde{\pi}_t(x)]) \Big],
\end{align*}
where $\mathbb{E}_x[\pi_t(x)] \approx \mathbb{E}_x[\tilde{\pi}_t(x)] = {\textbf{w}^N_t}^\top \textbf{X}^N_t$ is the barycenter of $\pi_t$, MV and MD correspond to the convergence of $\pi_t$, with a smaller value indicating convergence to the global maximum.

We compared these two metrics against BR and simple regret, $f(x^*_\text{true}) - \max_{x \in \textbf{X}_t} f(x)$. Experiments were conducted using the Ackley (see Table \ref{tab:exp}) over six iterations with 20 repeats (120 data points). Firstly, the left side in Figure \ref{fig:corr} shows a similar convergence trend for SR, BR, and MV, particularly noting that SOBER-LFI converges surprisingly quickly. The linear correlation matrix on the right implies that both MD and MV are highly correlated with BR, clearly explaining that $\pi_t$ convergence in Eq.(\ref{eq:dual}) is a good proxy for BR. $\pi_t$ (the MC estimate of $\hat{x}^*_t$) shrinks toward the true global maximum, $x^*_\text{true}$, with smaller variance (more confidence), and both are linearly correlated with BR minimisation.

One potential explanation for the significant performance improvement of SOBER-LFI is the synergy between the explorative KQ approach and the exploitative LFI synthetic likelihood. As illustrated in Figure \ref{fig:sober}, the LFI exhibits greater peakedness around the current maximum, $\hat{y}^*_t$, compared to the TS distribution. Such a distribution is likely to result in smaller MV. Our KQ method is capable of robustly selecting small areas of uncertainty, even with such a peaked distribution (refer back to Figure \ref{fig:sober}). In essence, the exploitative nature of LFI contributes to the reduction of MV, and consequently, to a decrease in BR, whereas the KQ facilitates robust exploration under peaked distribution.

\subsection{Robustness Analysis}
\begin{figure}
    \centering
    \includegraphics[width=0.9\textwidth]{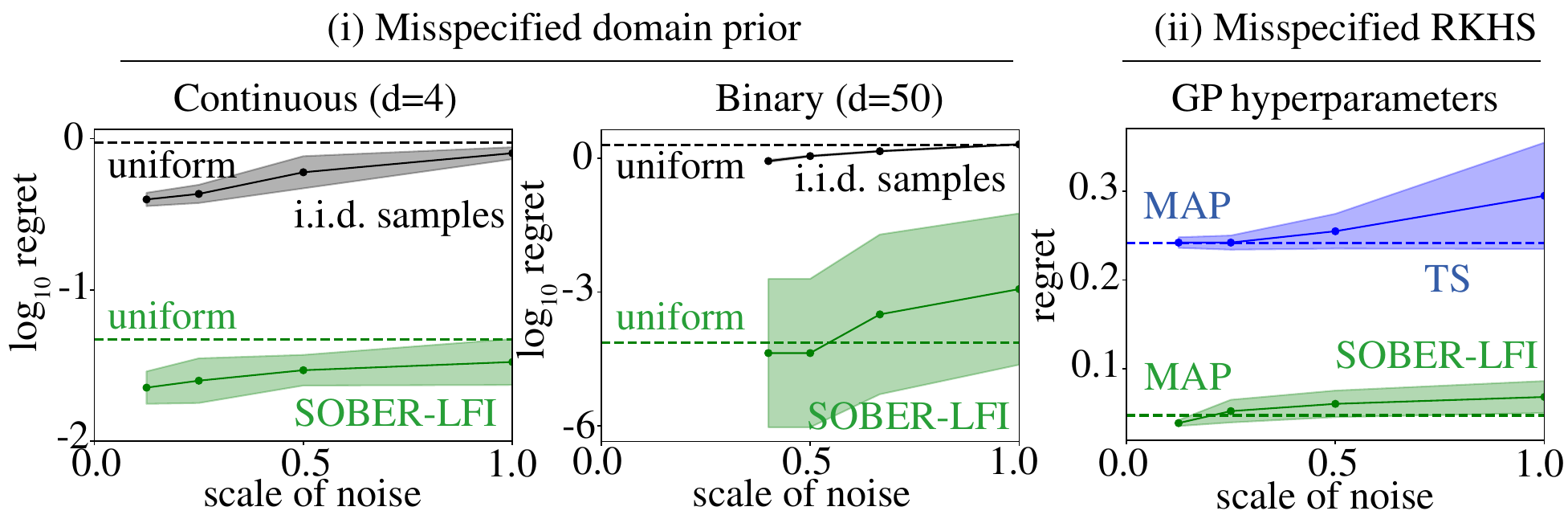}%
    \caption{Robustness analysis on Ackley ($n = 200$, $\log_{10}$ regret at 10th iteration) (i) misspecified domain prior: The left and middle experiments are examined misspecified domain prior for continuous and binary optimisation. (ii) Misspecified RKHS: We added noise to the GP hyperparameters that were tuned by MLE. In all misspecification cases, SOBER showed great resilience against misspecification noise.
    }
    \label{fig:robustness}
  \vspace{-1em}
\end{figure}
\paragraph{Misspecified Domain Prior.}
We evaluated the robustness of our approach to a misspecified domain prior, $\pi_0$, by introducing noise to the hyperparameters of $\pi_0$ as depicted in Figure \ref{fig:robustness}(i). Note that, although noise is added to the $\pi_0$ hyperparameters, they are quickly updated via SIR, suggesting that the system should be resilient to initial misspecifications. Hence, this experiment primarily assesses robustness against skewed initial sample configurations, considering we generate the initial 100 samples by drawing from $\pi_0$. For continuous optimisation, we employed a uniform distribution as the non-informative prior (reference) and a truncated Gaussian distribution as the misspecified (biased) prior. We maintained a fixed covariance matrix, $\textbf{I}_{d \times d}$, but introduced noise to the mean vector, $\mu_\pi = \sigma \epsilon$, where $\epsilon \sim \mathcal{U}(0, 1)$ represents uniform noise and $\sigma$ denotes the noise scale. In the case of binary optimisation, we used a Bernoulli distribution, with its probability vector treated as the stochastic variable $p = \sigma \epsilon$ ($p = 0.5$ denotes uniform). In both scenarios, as $\sigma$ approaches zero, the system converges to the global maximum, $x^*_\text{true} = [0]^d$, indicating that a smaller $\sigma$ favours the identification of the global maximum. The observed simple regrets remain nearly constant across different noise scales and significantly outperform i.i.d. batch samples drawn from the domain prior $\pi_0$.

\paragraph{Misspecified RKHS.}
We investigated the robustness against a misspecified RKHS, specifically misfit GP hyperparameters, as shown in Figure \ref{fig:robustness}(ii). Referring to §\ref{sec:misspecification}, the worst-case error estimate in Proposition \ref{prop:robustness} is guaranteed to be uniformly bounded. Noise was introduced to the GP hyperparameters, which were initially optimised using type-II MLE with the BoTorch optimiser \citep{balandat2020botorch}. The hyperparameters were adjusted as $\theta := \theta_\text{MLE} (1 + \sigma_\theta \epsilon)$, where $\epsilon \sim \mathcal{U}(-0.5, 0.5)$. The dashed lines represent the scenarios without noise (MAP cases). While the regret associated with batch TS \citep{kandasamy2015high} worsened and exhibited greater variance with increasing noise scale, SOBER-LFI achieved a plateau, indicating uniform robustness against the worst-case error. This demonstrates the susceptibility of TS to model misspecification, as exemplified in Figure \ref{fig:demo}.

\subsection{Scalability Analysis}\label{sec:scalability}
\begin{figure}
    \centering
    \subfloat[\centering dimensional scalability]{{\includegraphics[width=0.45\hsize]{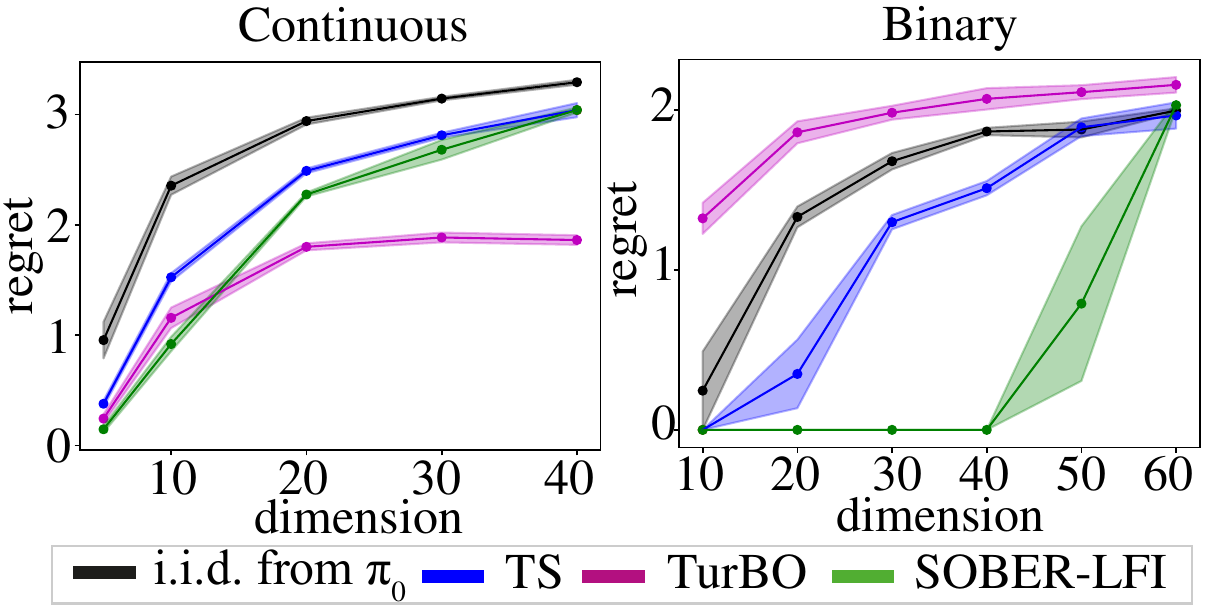} }}%
    \qquad
    \subfloat[\centering Computational complexity]{{\includegraphics[width=0.48\hsize]{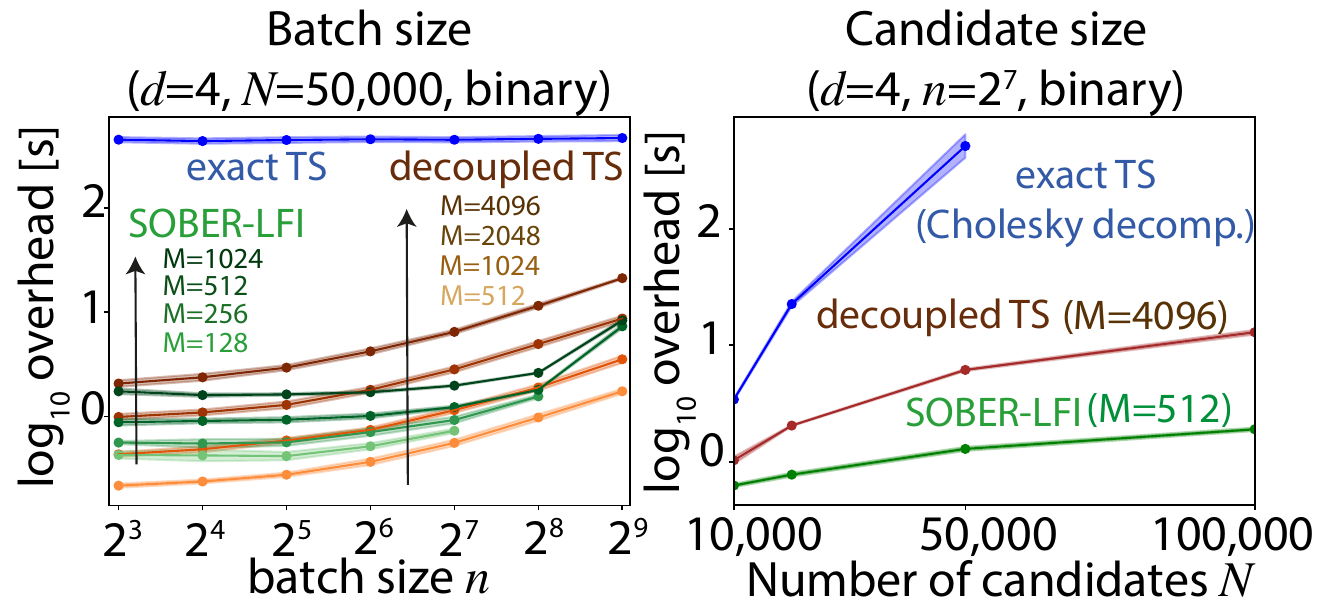} }}%
    \caption{Scalability Analysis: 
    (a) The regret over a varied number of dimensions of noisy Ackley function on both continuous and binary domains (batch size $n = 200$, and regrets are at 10th iterations. 
    (b). The overhead time over varied batch size $n$ and the number of candidates $N$. 
    The number of function samples $M$ approximates original kernel, which corresponds to number of samples for Nyström approximation in SOBER, and the ones for random Fourier features (RFF) approximation in decoupled TS.
    }%
  \label{fig:saclability}
\end{figure}
\paragraph{Dimensional Scalability.}
We assessed dimensional scalability by comparing our method against TurBO \citep{eriksson2019scalable}, a widely recognised method for high-dimensional BO. Since TurBO is designed solely for continuous domains, we adapted its algorithm by thresholding (recall §\ref{sec:experiments}). In the continuous domain, as shown in Figure \ref{fig:saclability}(a), while TurBO exhibits superior performance in dimensions exceeding 15, SOBER-LFI is more effective in lower dimensions. In binary optimisation, SOBER-LFI surpasses all baselines, even in 60 dimensions. This is because the binary space has fewer potential candidates, $2^d$, than the continuous space, allowing the hypercontractivity of the random convex hull to ensure that the empirical measure, $\tilde{\pi}_t$, adequately spans the entire domain.

\paragraph{Computational Complexity.}
We evaluated the computational overhead for batch queries. As detailed in §\ref{sec:inpractice}, the complexity is $\mathcal{O}(C_\varphi N + n^3 \log(N/n))$. At first glance, the cubic term related to batch size, $n$, appears unscalable; however, it is actually competitive, compared to TS and its variants. This is attributed to our candidate size, $N \gg n$, which makes the linearity in $N$ more impactful than the cubic term in $n^3$ for practical applications. Figure \ref{fig:saclability}(b) illustrates that SOBER-LFI significantly outpaces exact TS, which relies on Cholesky decomposition with a complexity of $\mathcal{O}(N^3)$. The fast variant using random Fourier features (decoupled TS, \cite{wilson2020efficiently}) achieves linearity with $N$, similar to our approach. However, it incurs a larger approximation error than the Nyström approximation \citep{yang2012nystrom}, necessitating more function samples, $M$, than Nyström. As Figure \ref{fig:saclability}(b) shows, within the practical range of parameter sets ($N = 20,000$, $M=512$, $n \leq 2^9$), our SOBER-LFI achieves fast computation. Yet, the cubic term, $n^3$, escalates quickly for $n > 1,000$, limiting scalability to such batch sizes. Nevertheless, considering the maximum batch sizes used in high-throughput drug discovery are typically 384 compounds \citep{carpentier2016hepatic}, we argue that SOBER-LFI remains sufficiently scalable for practical applications.

\subsection{Hyperparameter Sensitivity}
\begin{figure}
  \centering
  \includegraphics[width=1\hsize]{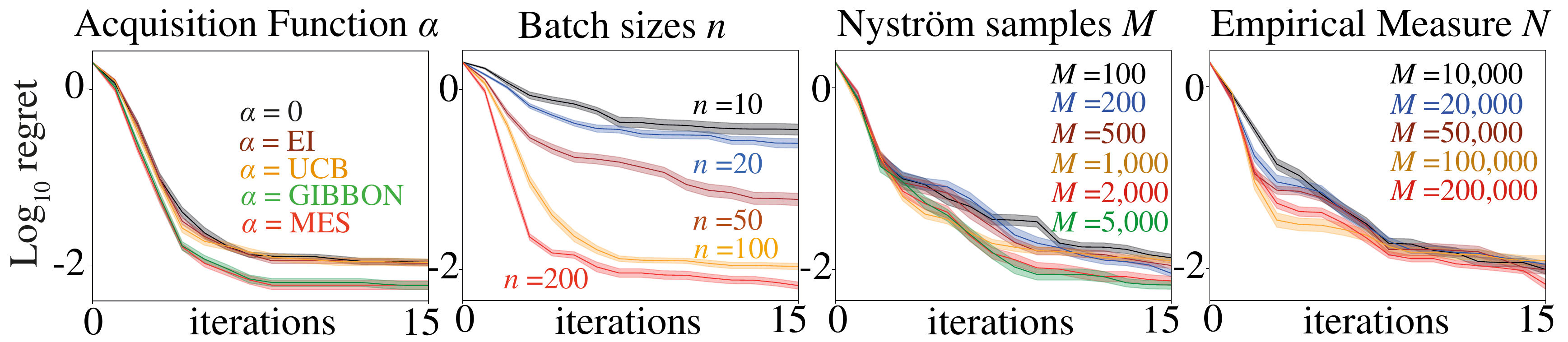}
  \caption{Hyperparameter sensitivity analysis using the Ackley function.}
  \label{fig:hyp}
  \vspace{-1em}
\end{figure}
The hyperparameter sensitivity of SOBER-LFI was examined using the Ackley, focusing on the effects of AFs ($\alpha$), batch size ($n$), the number of Nyström samples ($M$), and empirical measure sizes ($N$). The baseline conditions were set to $n = 100$, $\alpha = 0$ (no acquisition functions as reward), $M = 500$, and $N = 20,000$. For AFs, the information-theoretic AFs significantly enhances the convergence rate, whereas UCB/EI show no substantial change. This can be because MES function gives more dissimilar reward function shape to $\pi$ (= PI) than EI/UCB. Notably, our SOBER library integrates seamlessly with the popular GPyTorch/BoTorch library, allowing users to directly use AF and kernel instances defined by these libraries. More details can be found in our GitHub tutorials.

Regarding batch size ($n$), we observed an improvement in the convergence rate proportional to the batch size. Although this increase seems intuitive, it is not consistently observed in existing baselines (e.g., see Figure 6 in GIBBON \citep{moss2021gibbon}). For the Nyström samples ($M$) and empirical measure sizes ($N$), larger sample numbers correspond to faster convergence, reflecting our bound in Theorem \ref{thm:plain}. Specifically, a larger $M$ reduces the Nyström approximation error (the first term), and a larger $N$ decreases the empirical measure approximation error (the second term). This straightforward relationship between MC estimate error and sample size is not always evident in existing baselines (e.g., max-value entropy search, MES; \citet{wang2017max},  see §2.3 in \cite{takeno2023posterior}). However, an increase in sample numbers also results in higher computational overhead, as discussed in §\ref{sec:scalability}. Our default settings are competitive for real-world experiments discussed later, though they can be adjusted to balance the cost of queries \citep{adachi2022fast}\footnote{See guidelines in Appendix 2.2.2 in \citet{adachi2022fast}}.

\subsection{Adaptability Analysis}\label{sec:adabatch_exp}
\begin{figure}
    \centering
    \subfloat[\centering Adaptive batch sizes on Hartmann ($d=6$)]{{\includegraphics[width=0.45\hsize]{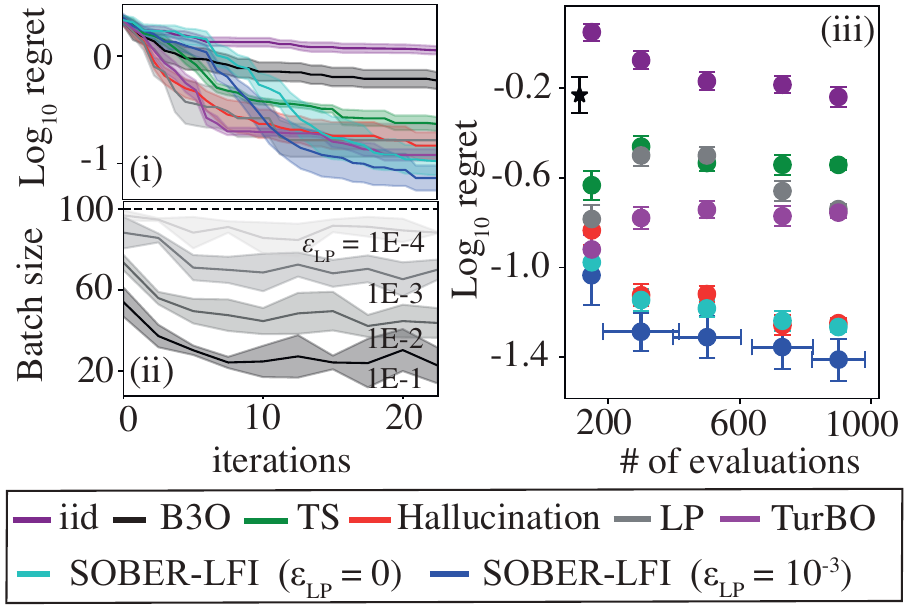} }}%
    \qquad
    \subfloat[\centering Adaptive safe exploration on Branin ($d=2$)]{{\includegraphics[width=0.48\hsize]{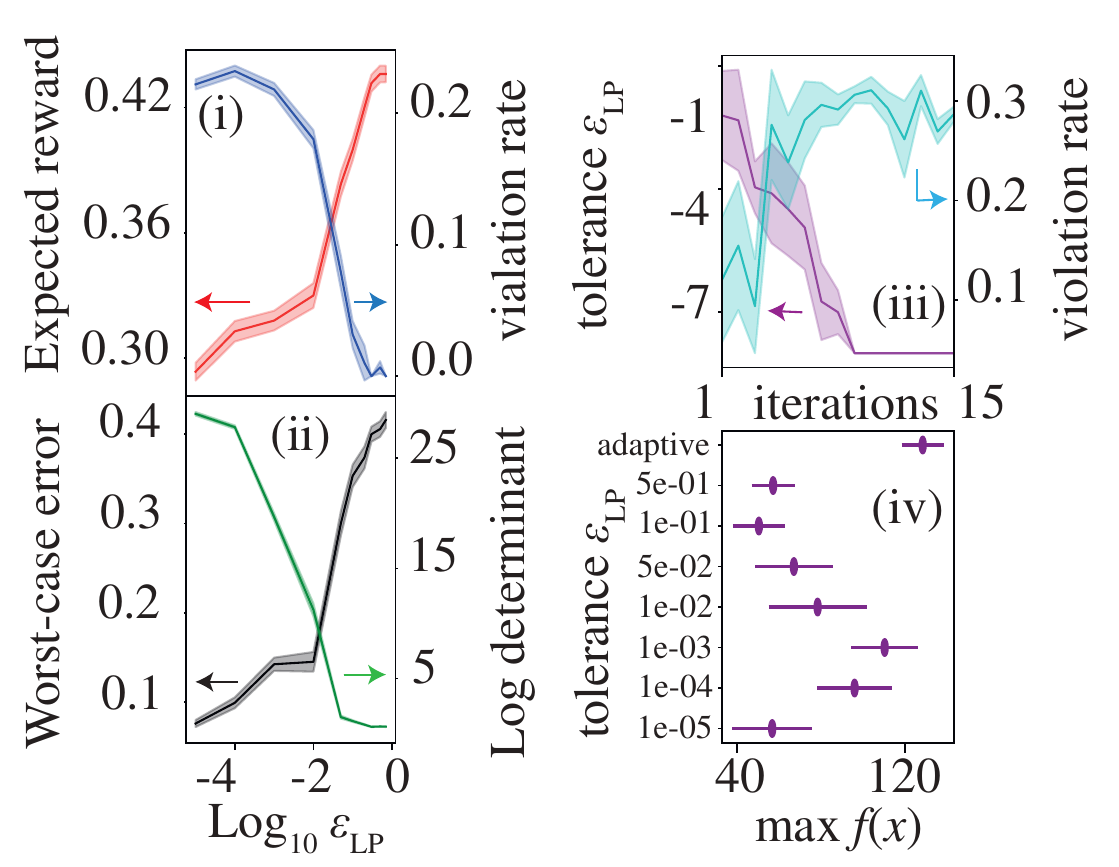} }}%
    \caption{Adaptability Analysis: 
    (a)(i) convergence plot with ($n \leq 5$). (ii) batch size variability ($n \leq 100$). The tolerance is set ($\epsilon_\text{LP} = 10^{-1}, 10^{-2}, 10^{-3}, 10^{-4}$). (iii) Total queries vs. simple regret at the last iteration results of (i)(ii). For fixed batch size baselines, the mean batch size of SOBER-LFI with $\epsilon_\text{LP} = 10^{-2}$ is used ($n = 5, 30, 50, 73, 90$).
    (b) Tolerance effect on constrained batch BO: the balance between (i) violation rate and expected reward, and (ii) worst-case error and log determinant. (iii) Tolerance adaptively controls violation rate, and (iv) outperforms the fixed cases. (i)(ii)(iii) are the two Y-axis plots where the colour and arrow indicate which Y axis to see.
    }%
    \label{fig:tolerance}
  \vspace{-1em}
\end{figure}

Within this section, we have explored the $\epsilon_\text{LP} > 0$ setting described in §\ref{sec:LP_constrained}.
\paragraph{Adaptive Batch Sizes.}
Our initial investigation focused on the effectiveness of adaptive batch sizes. To facilitate a comparison with B3O \citep{nguyen2016budgeted}, the only other baseline method employing adaptive batch sizes, we examined batch BO \emph{without} unknown constraints, i.e., $\tilde{q}(x) = 1$, albeit with $\epsilon_\text{LP} > 0$. As shown in Figure \ref{fig:tolerance}(a), SOBER-LFI, employing adaptive batch sizes, consistently outperformed the baseline methods across all experiments. An increase in $\epsilon_\text{LP}$ led to a reduction in batch size, which consistently decreased over iterations for all values of $\epsilon_\text{LP}$. This pattern indicates that SOBER-LFI initially requires a larger number of exploratory samples before narrowing its search space for exploitation. When compared to methods using fixed batch sizes within the same total cost framework, SOBER-LFI achieved lower regret, surpassing even the original SOBER with fixed batch sizes. Unlike B3O, which tends towards small batch sizes (around 4), the batch size of SOBER-LFI is tunable based on $n$ and $\epsilon_\text{LP}$.

\paragraph{Adaptive Safe Exploration.}
We then examined the influence of the expected violation rate, $\epsilon_\text{vio}$, in constrained BO, treating it as a time-varying tolerance, $\epsilon_\text{vio} = \epsilon_\text{LP}$ with $\epsilon_\text{LP} > 0$. The main findings are depicted in Figure \ref{fig:tolerance}(b).
\begin{tcolorbox}[colback=white]
\textbf{Four key metrics:}
\begin{compactenum}
    \item[(1)] \emph{The expected reward} (LP objective): A proxy for the safety level of our exploration strategies.
    \item[(2)] \emph{The violation rate} $1 - \lvert \tilde{\textbf{X}}_\text{batch} \rvert / \lvert \textbf{X}_\text{batch} \rvert$: A proxy for the actual safety level achieved during exploration.
    \item[(3)] \emph{The worst-case error} $\text{wce}[Q_{\pi_t, C_{t-1}}(\lvert \tilde{\textbf{X}}_\text{batch} \rvert)]$: the precision of quadrature.
    \item[(4)] \emph{log determinant} $\log \lvert  K(\tilde{\textbf{X}}_\text{batch}, \tilde{\textbf{X}}_\text{batch}) \rvert$: a proxy for the batch sample diversity.
\end{compactenum}
\end{tcolorbox}

We evaluated the impact of $\epsilon_{\text{LP}}$ on these metrics, aligning $\epsilon_{\text{LP}}$ with $\epsilon_{\text{vio}}$ to enable adaptive exploration in line with specified risk levels, $\epsilon_{\text{vio}}$, illustrated in Figures \ref{fig:tolerance}(b)(i)(ii). As risk levels increase, ensuring safety becomes a priority, leading to an uptick in the expected reward and a corresponding reduction in the violation rate, signifying safer exploration practices. Numerically, a higher risk level correlates with an increased worst-case error, indicating a relaxation in precision requirements, and a reduced log determinant, suggesting a decrease in the diversity of batch samples due to the proximity of selected points $(\textbf{X}^N_t)$ to each other. Conversely, lower risk levels favour a more optimistic and exploratory approach. Our results affirm that setting $\epsilon_{\text{LP}}$ equal to $\epsilon_{\text{vio}}$ allows our batch exploration strategy to adeptly adjust to varying risk levels. 

Additionally, we observed the evolution of the expected violation rate, $\epsilon_\text{vio}$, throughout the optimisation process. As depicted in Figure \ref{fig:tolerance}(b)(iii), $\epsilon_\text{vio}$ starts high and gradually decreases, highlighting an initial focus on safe data collection before shifting towards broader exploration. This strategy is reminiscent of `safe' BO approaches like those proposed by \citep{sui2015safe}, which have shown strong empirical performance and theoretical support (e.g., Figure 4 in \citet{xu2023constrained}). The inherent adaptability of our method to adjust batch sizes and tolerance levels showcases its efficiency, particularly as demonstrated by the more fast convergence compared to fixed tolerance approaches in Figure \ref{fig:tolerance}(b)(iv). Interestingly, the most effective fixed tolerance was $\epsilon_\text{LP} = 10^{-3}$, indicating that SOBER-LFI surpasses the performance of the exact case ($\epsilon_\text{LP} = 0$) under constraints, even with a fixed tolerance.

\subsection{Discrete and Mixed Variable Experiments}
\begin{figure}
  \centering
  \includegraphics[width=0.95\hsize]{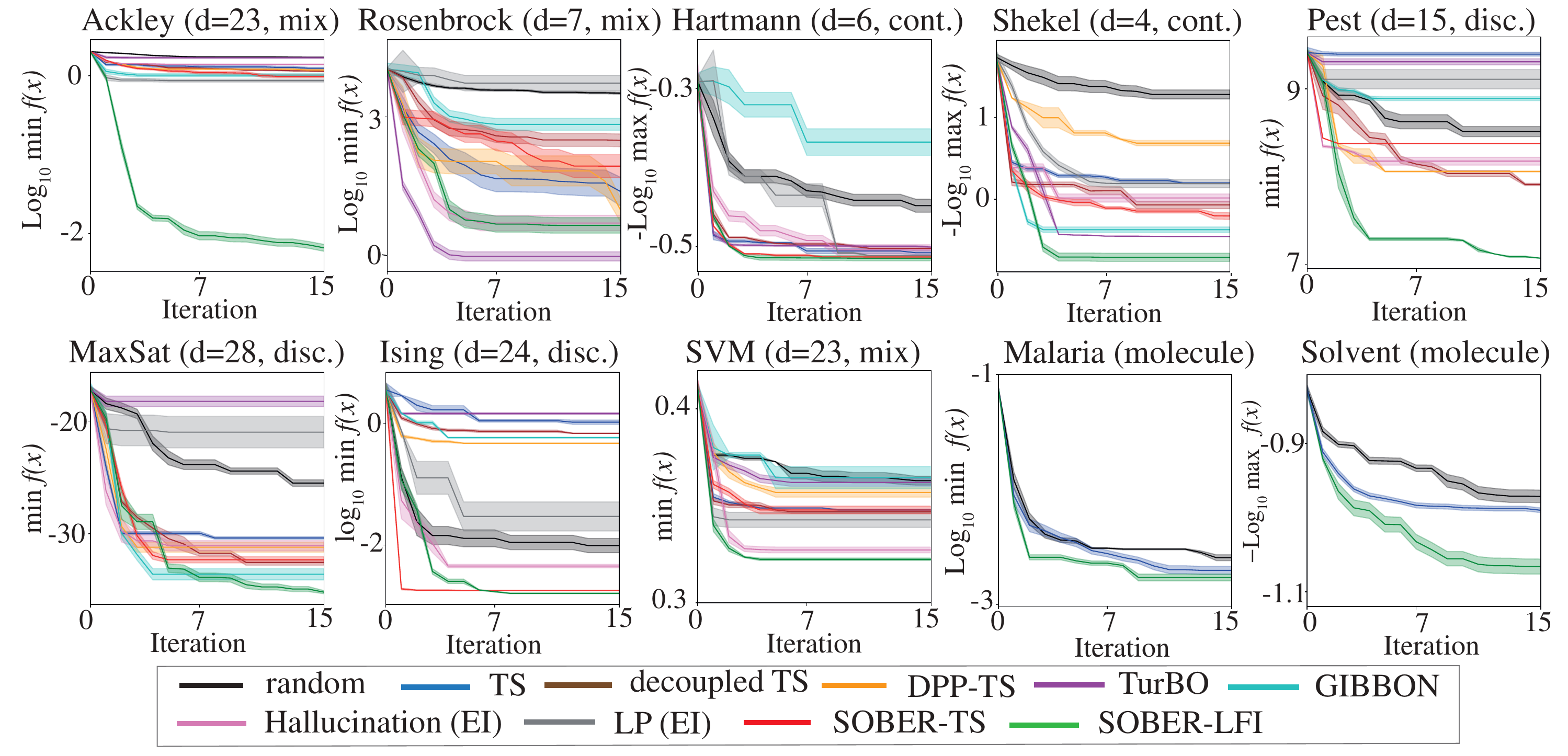}
  \caption{Tolerance effect on constrained batch BO on Branin ($d=2$): the balance between (a) violation rate and expected reward, and (b) worst-case error and log determinant. (c) Tolerance adaptively controls violation rate, and (d) outperforms the fixed cases. (a)(b)(c) are the two Y-axis plots where the colour and arrow indicate which Y axis to see.}
  \label{fig:regret_bo}
\end{figure}
\begin{table}[ht]
    \small
    \centering
    \resizebox{1\textwidth}{!}{
    \begin{tabular}{lcccccccccc|c}
    \toprule
    baselines & 
    Ackley &
    Rosenbrock &
    Hartmann &
    Shekel &
    Pest &
    MaxSat &
    Ising & SVM & 
    Malaria &
    Solvent &
    Mean rank\\
    \midrule
    Random & -1.92	& -1.96 & -1.26 & -1.17 & -1.92 & -1.89 & -1.64 & 0.82 & 1.40 & 1.49 &
    -
    \\
    batch TS &
    2.71	& 3.10 & 2.79 & 2.86 &	3.00 &	3.70 & 3.22 & 3.36 & 2.71 & 2.85 &
    3.1
    \\
    decoupled TS &
    \textbf{2.20}	& \textbf{2.04} & \textbf{2.01} & \textbf{2.04} & 3.17 & 3.22 & 3.65 & 3.90 & - & - &
    2.6
    \\
    DPP-TS &
    4.85	& 4.56 & 4.35 & 4.62 & 5.67 & 4.49 & 4.73 & 4.73 & - & - &
    7.4
    \\
    TurBO &
    3.42 &	3.06 & 2.12 & 3.07 &	\textbf{2.91} & 2.97 & 3.45 & 3.58 & - & - &
    3.3
    \\
    GIBBON &
    4.92 & 4.18 & 3.71 & 3.52 & 3.72 & 4.71 & 4.25 & 4.41 &  - & - &
    6.8
    \\
    Hallucination &
    4.52 &	4.09 & 4.42 & 3.68 & 4.68 & 4.75 & 4.14 & 5.05 & - & - & 
    7.4
    \\
    LP &
    5.50 &	5.48 & 5.23 & 4.78 & 3.84 & 5.48 & 5.10 & 4.53 & - & - &
    8.5
    \\
    SOBER-TS &
    3.10 & 3.43 & 3.16 & 3.17 & 3.30 & 4.01 & 3.20 & 3.21 & - & - &
    4.1
    \\
    SOBER-LFI &
    2.58 & 2.19 & 2.08 & 2.65 & 2.99 & \textbf{2.96} & \textbf{2.28} & \textbf{2.31} & \textbf{2.43} & \textbf{2.35} & \textbf{1.5} \\
    \bottomrule
    \end{tabular}
    }
    \caption{Average cumulative wall-clock time for 15 iterations (log10 second).}
    \label{tab:time}
    \vspace{-1em}
\end{table}

Figure \ref{fig:regret_bo} and Table \ref{tab:time} showcase the convergence performance and the wall-clock time for sampling overhead at the 15th iteration, respectively. SOBER-LFI surpasses nine baselines in nine out of ten experiments, demonstrating its versatility and effectiveness across a wide range of multimodal and noisy functions in continuous, discrete, and mixed spaces. Although SOBER-LFI did not achieve the top performance on the unimodal Rosenbrock function---which tends to favour more exploitative algorithms like TurBO---it secured a strong second place. This performance underscores the efficiency of SOBER-LFI's strategy in dynamically narrowing the sampling region around the global maximum. In the realm of drug discovery, SOBER-LFI distinguished itself by showing fast convergence, areas where most algorithms falter due to specific kernel and space requirements. The solvent data set, in particular, highlights scenarios where batch TS quickly converges in early stages but fails to escape local maxima, eventually equating its final regret with that of random search. Conversely, SOBER-LFI avoids such pitfalls via exploratory KQ sampling.

\subsection{Constrained Optimisation Experiments}
\begin{figure}
  \centering
  \includegraphics[width=1\hsize]{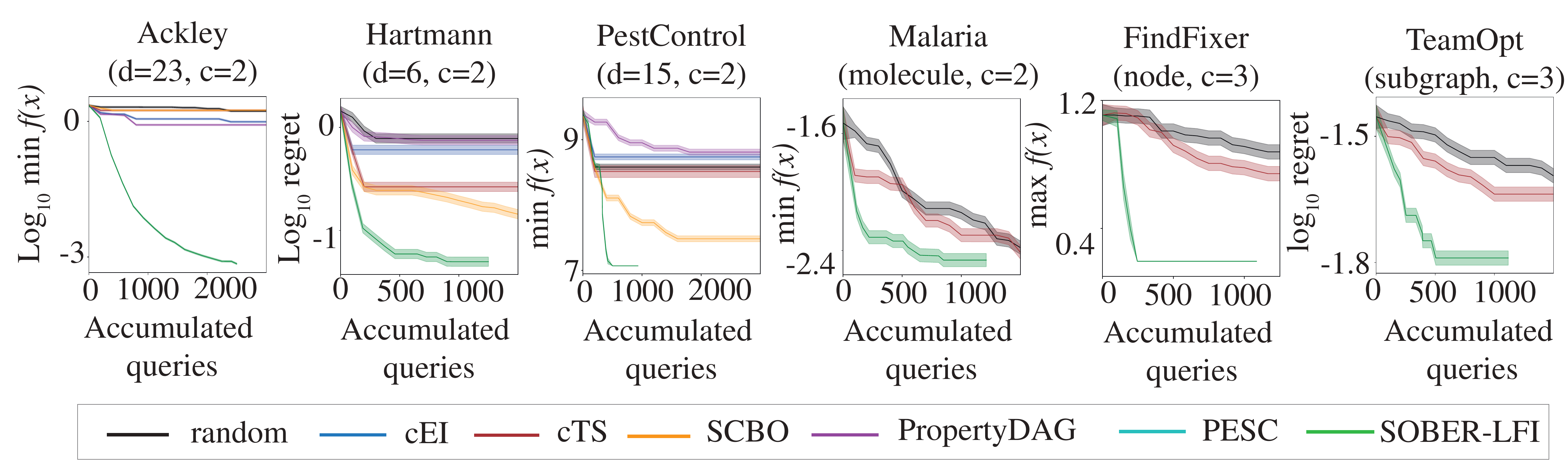}
  \caption{Convergence plot of constrained batch Bayesian optimisation results. $d$ is the dimension, $c$ is the number of unknown constraints.}
  \label{fig:regret_cbo}
\end{figure}
In the domain of constrained BO tasks with adaptive batch sizes, SOBER-LFI stands out as the sole method offering adaptive batching under constraints. We defined the upper limit of batch sizes for comparison (as detailed in Table \ref{tab:exp}). While baseline methods maintain fixed upper bound batch sizes across iterations, SOBER-LFI adeptly adjusts its batch sizes, leading to a more efficient use of queries. Consequently, SOBER-LFI typically requires fewer queries to reach the same iteration $T$. Figure \ref{fig:regret_cbo} highlights SOBER-LFI's robust empirical performance across various tasks.

\subsection{Simulation-based Inference Experiments}
\begin{figure}
  \centering
  \includegraphics[width=0.9\hsize]{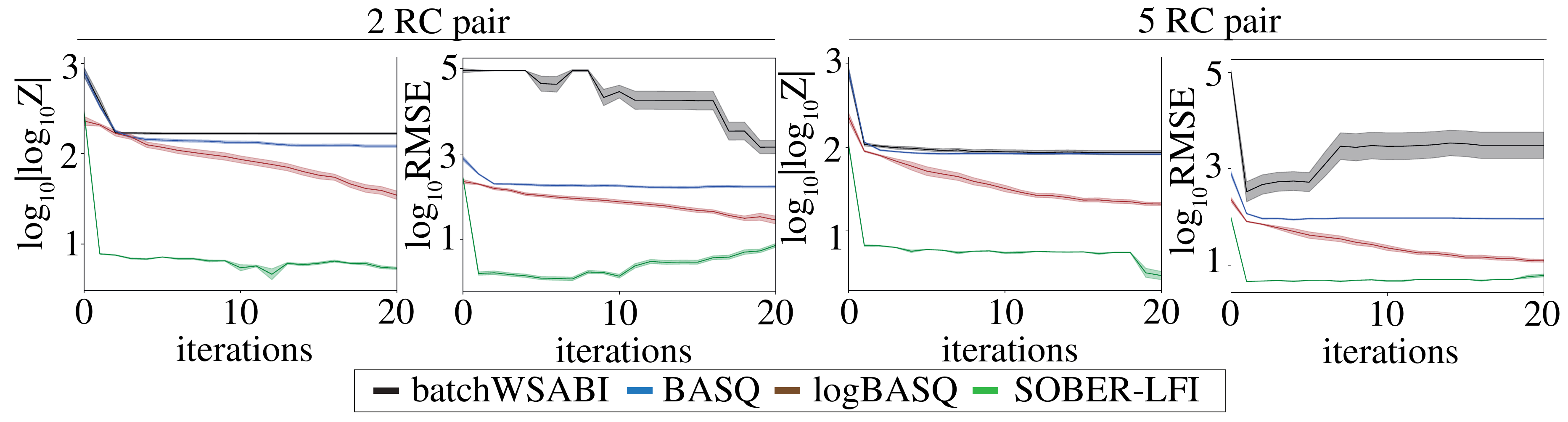}
  \caption{Batch Bayesian Quadrature baseline comparisons across two real-world simulation-based inference tasks. These tasks evaluate both approximation error of evidence ($Z$) and posterior approximation as root-mean-squared-error (RMSE) against true posterior distribution via exhaustive MCMC sampling and its kernel density estimation. Lower is better for both metrics.}
  \label{fig:regret_bq}
  \vspace{-1em}
\end{figure}
SOBER-LFI also demonstrates superior performance against batch BQ baselines in simulation-based inference tasks, as documented by \citet{adachi2023bayesian}. In these tasks, the prior variance significantly exceeds that of the likelihood, resulting in a sharply peaked posterior distribution. This condition renders exploration of the prior's tail as excessive. While the original BASQ method tends toward over-exploration of the prior distribution, leading to performance plateaus, the logBASQ variant mitigates this issue through log-warp modeling. Nevertheless, SOBER-LFI outpaces all baselines by a significant margin in both posterior and evidence inference across all tasks. It achieves this by effectively concentrating $\pi$ towards the posterior mode, thereby circumventing unnecessary over-exploration.

\section{Discussion}\label{sec:discussion}
We introduced \emph{SOBER}, a novel quadrature approach to batch BO through probabilistic lifting, showcasing its versatility and theoretical robustness. Initially, we elucidated the theoretical underpinnings of SOBER as a KQ method, emphasizing that its worst-case error bound predominantly stems from the Nyström approximation and empirical measure approximation errors. Crucially, SOBER maintains bounded errors even in scenarios where the RKHS is misspecified, thereby ensuring robustness against GP misspecification. The method's incorporation of closed-form LFI synthetic likelihood, SIR, and recombination techniques facilitates versatile, fast, and diverse sampling strategies. Empirical evidence further verifies that the diminishing variance of $\pi_t$ correlates strongly with BR, suggesting that the synergistic effect of the exploitative LFI $\pi_t$ and the explorative KQ sampling constitutes an effective heuristic for addressing the probabilistically lifted dual objective presented in Eq.(\ref{eq:dual}). Through extensive empirical analysis, we demonstrated SOBER-LFI's robustness, scalability, insensitivity to hyperparameters, and adaptivity, further bolstered by an extensive comparative study across a broad spectrum of real-world applications.

Despite its promising empirical performance, there is a pressing need for deeper theoretical analysis to enhance our understanding and identify avenues for improvement. The practical superiority of subsample-based KQ, relative to theoretical predictions, remains a puzzle. Theories on random convex hulls and hypercontractivity provide promising insights, although these explanations originate from slightly different contexts. Additionally, the convergence rate of the KQ method employed is limited to the non-adaptive case (single batch selection). This limitation arises because KQ is based on assumptions of a more general probability measure, such as a non-compact domain. However, \citet{kanagawa2019convergence} established the convergence rate for sequential adaptive BQ within a compact domain. In BO tasks, the assumption of a compact domain is essential (without it, convergence cannot be guaranteed), indicating that this area of research could potentially lead to establishing a full convergence rate.
Additionally, exploring the implications of SOBER-LFI in the context of Bayesian cumulative regret convergence rates is pivotal for unraveling the mechanisms behind its empirical success. However, we wish to highlight that our addressed issues---misspecified RKHS for general AFs, and adaptive batch sizes for a limited budget---are not yet accommodated by regret convergence analysis in the existing theoretical BO literature (refer to §\ref{sec:adabatch} and §\ref{sec:misspecification}). Our extensive empirical study, along with the theoretical bounds we present for worst-case errors, could lay the groundwork for enhanced theoretical insights.
It is also worth noting that SOBER currently does not accommodate asynchronous batch settings \citep{kandasamy2018parallelised}, a limitation identified in previous works \citep{adachi2022fast, adachi2024adaptive}. However, given its generality and flexibility, integrating SOBER with other advanced methods represents a fertile direction for future research, promising valuable contributions to the field for both practitioners and theoreticians. 

This paper is the journal extension of the non-archival ICML workshop paper \citep{adachi2023sober} and AISTATS paper \citep{adachi2024adaptive}. Although the ICML workshop paper \citep{adachi2023sober} has not undergone rigorous peer review, it presents content similar to our current work but includes only a limited number of experiments and no theories. The current paper builds upon this ICML workshop paper, providing an in-depth discussion on how to connect batch Bayesian optimization and Bayesian/kernel quadrature through a probabilistic lifting technique. In AISTATS paper \citep{adachi2024adaptive}, we focused on adaptivity, we posited that the link between quadrature and optimization was established in an ICML workshop paper \citep{adachi2023sober}. 


\acks{We thank Leo Klarner for the insightful discussion of Bayesian optimistaion for drug discovery, Samuel Daulton, Binxin Ru, and Xingchen Wan for the insightful discussion of Bayesian optimisation for graph and mixed space, Yannick Kuhn for preparing PyPI. Masaki Adachi was supported by the Clarendon Fund, the Oxford Kobe Scholarship, the Watanabe Foundation, and Toyota Motor Corporation. Satoshi Hayakawa was supported by the Clarendon Fund, the Oxford Kobe Scholarship, and the Toyota Riken Overseas Scholarship. Harald Oberhauser was supported by the DataSig Program [EP/S026347/1], the Hong Kong Innovation and Technology Commission (InnoHK Project CIMDA), and the Oxford-Man Institute. Martin Jørgensen was partly supported by the Research Council of Finland (grant 356498) and the Carlsberg foundation. Saad Hamid is grateful for funding from the Engineering and Physical Sciences Research Council of the UK.}


\appendix
\section{Proof of Proposition~\ref{prop:lp}} \label{app:proof}
\begin{proof}[Proof of Proposition~\ref{prop:lp}]
    Note that the constraint $\lvert \textbf{w}\rvert_0 \le n$ is automatically satisfied when we use the simplex method or its variant.
    Without this constraint, we have a trivial feasible solution $\textbf{w}=\textbf{w}^N_t$,
    so, for the optimal solution $\textbf{w}_*$, we have
    $\textbf{w}_*^\top \bigl[ \alpha_t(\textbf{X}^n_t) \odot \tilde{q}_t(\textbf{X}^n_t)\bigr]
        \ge {\textbf{w}^n_t}^\top \bigl[ \alpha_t(\textbf{X}^N_t) \odot \tilde{q}_t(\textbf{X}^N_t)\bigr]$.
    Since $\mathbb{E}[{\tilde{\textbf{w}}^n_t}{}^\top \alpha_t(\tilde{\textbf{X}}^n_t)] = {\textbf{w}^n_t}^\top \bigl[ \alpha_t(\textbf{X}^n_t) \odot \tilde{q}_t(\textbf{X}^n_t)\bigr]
    = \textbf{w}_*^\top \bigl[ \alpha_t(\textbf{X}^N_t) \odot \tilde{q}_t(\textbf{X}^N_t)\bigr]$,
    we obtain the first estimate Eq.~(\ref{eq:lp-1}).

    For the latter estimate, we first decompose the error into
    two parts:
    \begin{align}
        &\mathbb{E}\!\left[\left\lvert {\tilde{\textbf{w}}^n_t}{}^\top
        f_{t-1}(\tilde{\textbf{X}}^n_t) -
        {\textbf{w}^N_t}^\top f_{t-1}(\textbf{X}^N_t)\right\rvert\right]
        \nonumber\\
        &\le
        \mathbb{E}\!\left[\left\lvert {\tilde{\textbf{w}}^n_t}{}^\top
        f_{t-1}(\tilde{\textbf{X}}^n_t) -
        {\textbf{w}^n_t}^\top f_{t-1}(\textbf{X}^n_t)\right\rvert\right]
        +
        \left\lvert {\textbf{w}^n_t}^\top
        f_{t-1}(\textbf{X}^n_t) -
        {\textbf{w}^N_t}^\top f_{t-1}(\textbf{X}^N_t)\right\rvert
        \label{eq:dcp}.
    \end{align}
    For the first term, considering each $x\in \textbf{X}^n_t$
    on whether or not it gets included in $\tilde{\textbf{X}}^n_t$,
    we have
    \begin{align*}
        &\mathbb{E}\!\left[\left\lvert {\tilde{\textbf{w}}^n_t}{}^\top
        f_{t-1}(\tilde{\textbf{X}}^n_t) -
        {\textbf{w}^n_t}^\top f_{t-1}(\textbf{X}^n_t)\right\rvert\right]
        \\&\le {\textbf{w}^n_t}^\top
        \Bigl[ \lvert f_{t-1} \rvert(\textbf{X}^n_t) \odot (1-\tilde{q}_t(\textbf{X}^n_t))\Bigr]
        \le {\textbf{w}^n_t}^\top (1-\tilde{q}_t(\textbf{X}^n_t))
        \max_{x\in\textbf{X}^n_t} \lvert f_{t-1}(x)\rvert
        \\&=\Bigl[1 - {\textbf{w}^n_t}^\top \tilde{q}_t(\textbf{X}^n_t)\Bigr]
        \max_{x\in\textbf{X}^n_t} \lvert f_{t-1}(x) \rvert
        \le \Bigl[1 - {\textbf{w}^N_t}^\top \tilde{q}_t(\textbf{X}^N_t)\Bigr]
        \max_{x\in\textbf{X}^n_t} \lvert f_{t-1}(x) \rvert,
    \end{align*}
    where the last inequality follows from the inequality constraint
    $(\textbf{w} - \textbf{w}^N_t)^\top \tilde{q}_t(\textbf{X}^N_t) \ge 0$
    in the LP.
    Since $\lvert f_{t-1}(x)\rvert=\lvert\langle f_{t-1}, C_{t-1}(\cdot, x) \rangle\rvert
    \le \lVert f_{t-1}\rVert C_{t-1}(x, x)^{1/2}$ from the reproducing property of RKHS,
    we obtain
    \begin{equation}
        \mathbb{E}\!\left[\left\lvert {\tilde{\textbf{w}}^n_t}{}^\top
        f_{t-1}(\tilde{\textbf{X}}^n_t) -
        {\textbf{w}^n_t}^\top f_{t-1}(\textbf{X}^n_t)\right\rvert\right]
        \le \epsilon_\text{vio}K_{\max}\lVert f_{t-1}\rVert.
        \label{eq:proof-lp-1}
    \end{equation}

    Let us then bound the second term of the RHS of Eq.~(\ref{eq:dcp}).
    Note that, from the formula of worst-case error of kernel quadrature
    (see, e.g., \citep[][Eq. (14)]{hayakawa2022positively}),
    we can bound
    \begin{align}
        \left\lvert {\textbf{w}^n_t}^\top
        f_{t-1}(\textbf{X}^n_t) -
        {\textbf{w}^N_t}^\top f_{t-1}(\textbf{X}^N_t)\right\rvert^2
        \le \lVert f_{t-1} \rVert^2
        (\textbf{w}_* - \textbf{w}^N_t)^\top
        C_{t-1}(\textbf{X}^N_t, \textbf{X}^N_t) (\textbf{w}_* - \textbf{w}^N_t)
        \label{eq:proof-lp-2}
    \end{align}
    (recall $\textbf{w}_*$ has the same dimension as $\textbf{w}^N_t$).
    We now want to estimate
    \begin{equation}
        (\textbf{w}_* - \textbf{w}^N_t)^\top
        C_{t-1}(\textbf{X}^N_t, \textbf{X}^N_t) (\textbf{w}_* - \textbf{w}^N_t).
        \nonumber
    \end{equation}
    Consider approximating $C_{t-1}$ by $\tilde{C}_{t-1}$.
    Since $C_{t-1} - \tilde{C}_{t-1}$ is positive semi-definite from the property
    of Nystr{\"o}m approximation (see, e.g., the proof of \citep[][Corollary 4]{hayakawa2022positively}),
    for any $x, y \in \textbf{X}^N_t$, we have
    \[
        \lvert (C_{t-1} - \tilde{C}_{t-1})(x, y)\rvert
        \le \lvert (C_{t-1} - \tilde{C}_{t-1})(x, x)\rvert^{1/2}
        \lvert (C_{t-1} - \tilde{C}_{t-1})(y, y)\rvert^{1/2}
        \le \epsilon_\text{nys}^2.
    \]
    Thus, we have
    \begin{align}
        &(\textbf{w}_* - \textbf{w}^N_t)^\top
        \Bigl[(C_{t-1} - \tilde{C}_{t-1})(\textbf{X}^N_t, \textbf{X}^N_t)\Bigr] (\textbf{w}_* - \textbf{w}^N_t)
        \nonumber\\
        &\le (\textbf{w}_* + \textbf{w}^N_t)^\top (\epsilon_\text{nys}^2\boldsymbol{1}\boldsymbol{1}^\top) (\textbf{w}_* + \textbf{w}^N_t) = 4\epsilon_\text{nys}^2.
        \label{eq:proof-lp-3}
    \end{align}
    Finally, we estimate
    \begin{align}
        &(\textbf{w}_* - \textbf{w}^N_t)^\top
        \tilde{C}_{t-1}(\textbf{X}^N_t, \textbf{X}^N_t) (\textbf{w}_* - \textbf{w}^N_t)
        \nonumber \\
        &=(\textbf{w}_* - \textbf{w}^N_t)^\top
        \sum_{j=1}^{n-2}
        \boldsymbol{1}_{\{\lambda_j>0\}}\lambda_j^{-1}
        \varphi_j(\textbf{X}^N_t)\varphi_j(\textbf{X}^N_t)^\top
        (\textbf{w}_* - \textbf{w}^N_t) \nonumber\\
        &=\sum_{j=1}^{n-2}
        \boldsymbol{1}_{\{\lambda_j>0\}}\lambda_j^{-1}
        \left[(\textbf{w}_* - \textbf{w}^N_t)^\top \varphi_j(\textbf{X}^N_t)\right]^2.
        \label{eq:proof-lp-4}
    \end{align}
    From the inequality constraint in the LP,
    we have $\lvert(\textbf{w}_* - \textbf{w}^N_t)^\top \varphi_j(\textbf{X}^N_t)\rvert \le \epsilon_\text{LP}
    \sqrt{\lambda_j/(n-2)}$,
    so that Eq.~(\ref{eq:proof-lp-4}) is further bounded as
    \begin{align}
        (\textbf{w}_* - \textbf{w}^N_t)^\top
        \tilde{C}_{t-1}(\textbf{X}^N_t, \textbf{X}^N_t) (\textbf{w}_* - \textbf{w}^N_t) \le \sum_{j=1}^{n-2}
        \boldsymbol{1}_{\{\lambda_j>0\}}\lambda_j^{-1}
        \epsilon_\text{LP}^2\frac{\lambda_j}{n-2} \le \epsilon_\text{LP}^2.
        \label{eq:proof-lp-5}
    \end{align}
    By adding the both sides of Eqs.~(\ref{eq:proof-lp-3}) and (\ref{eq:proof-lp-5}),
    we obtain
    \[
        (\textbf{w}_* - \textbf{w}^N_t)^\top
        C_{t-1}(\textbf{X}^N_t, \textbf{X}^N_t) (\textbf{w}_* - \textbf{w}^N_t)
        \le 4\epsilon_\text{nys}^2 + \epsilon_\text{LP}^2
        \le (2\epsilon_\text{nys} + \epsilon_\text{LP})^2.
    \]
    By applying this to Eq.~(\ref{eq:proof-lp-2}), we have
    $\left\lvert {\textbf{w}^n_t}^\top
        f_{t-1}(\textbf{X}^n_t) -
        {\textbf{w}^N_t}^\top f_{t-1}(\textbf{X}^N_t)\right\rvert
        \le \lVert f \rVert (2\epsilon_\text{nys} + \epsilon_\text{LP})$.
    Combining this with Eqs.~(\ref{eq:dcp}) and (\ref{eq:proof-lp-1}) yields
    the desired inequality Eq.~(\ref{eq:lp-2}).
\end{proof}
This proof is mirrored from our previous work \citep{adachi2024adaptive}.

\section{Experimental details}\label{app:exp}
Due to the page restriction to be within 35 pages on submission, we defer the details to our prior papers. While our previous papers and GitHub code work as the explanation to reproduce our results, we will update the experimental details to here in appendix together to be self-contained. The experimental details of batch constrained BO is delineated in our previous work \citep{adachi2024adaptive}. The batch BQ experiments are detailed in our previous work \citep{adachi2023bayesian}. For batch unconstrained BO, we used the constrained BO tasks without constraints. We explain the details of the rest; Rosenbrock, Shekel, MaxSat, Ising, SVM, and Solvent:

\paragraph{Synthetic: Rosenbrock function}
We modified the original Rosenbrock function \citep{simulationlib} into a 7-dimensional function with the mixed spaces of 1 continuous and 6 discrete variables, following \citet{daulton2022bayesian}. The first 1 dimension is continuous with bounds $[-4, 11]^1$. The other 6 dimensions are discretised to be categorical variables, with 4 possible values $x_1 \in \{-4, 1, 6, 11\}$. 

\paragraph{Synthetic: Shekel function}
We use Shekel function without any modification from \citep{simulationlib}, 4 dimensional continuous variables bounded $[0, 10]^4$.

\paragraph{Real-world: Maximum Satisifiability}
Maximum satisfiability (MaxSat in the main) is proposed in \citet{oh2019combinatorial}, which is 28 dimensional binary optimisation problem. The objective is to find boolean values that maximise the combined weighted satisfied clauses for the data set provided by Maximum Satisfiability competition 2018. Both code and data set are used in \url{https://github.com/xingchenwan/Casmopolitan} \citep{wan2021think}.

\paragraph{Real-world: Ising Model Sparsification}
Ising Model Sparsification (Ising in the main) is proposed in \citet{oh2019combinatorial}, which is 24 dimensional binary optimisation problem. The objective is to sparsify an Ising model using the regularised Kullback-Leibler divergence between a zero-field Ising model and the partition function, considering $4 \times 4$ grid of spins with regularisation coefficient $\lambda = 10^{-4}$. Code is in \url{https://github.com/QUVA-Lab/COMBO}.

\paragraph{Real-world: Support Vector Machine Feature Selection}
Support vector machine feature selection (SVM in the main) is proposed in \citet{daulton2022bayesian},  which is 23 dimensional mixed-type input optimisation problem (20 dimensional binary and 3 dimensional continuous variables). The objective is jointly performing feature selection (20 features) and hyperparameter optimisation (3 hyperparameters) for a support vector machine trained in the CTSlice UCI data set \citep{graf20112d, Dua2019}. Code is used in \url{https://github.com/facebookresearch/bo\_pr}.

\paragraph{Real-world: Polar solvent for batteries} The data set with 133,055 small molecules represented as 2048-dimensional binary features were optimised and predicted by the quantum-chemical computations, known as QM9 data set \citep{ramakrishnan2014quantum}. The target variable is the dipole moment, which is basically correlated with the solvation capability in electrolytes in lithium-ion batteries, increasing the ratio of electro-mobile lithium-ions. The higher the dipole moment becomes, the larger (better) the ionic conductivity does. The data set is downloaded from \url{http://quantum-machine.org/datasets/}. The coding was done with Gauche \citep{griffiths2024gauche}. Due to the low expressive capability of 2048-dimensional binary features, we removed the duplicated candidates that show identical binary features from the QM9 data set, then applied the batch BO experiments.

\vskip 0.2in
\bibliography{sample}

\end{document}